\documentclass[10pt,twocolumn,letterpaper]{article}

\usepackage[pagenumbers]{cvpr} 

\usepackage{graphicx}
\usepackage{amsmath}
\usepackage{amssymb}
\usepackage{booktabs}

\usepackage{enumitem}
\usepackage{times}
\usepackage{epsfig}
\usepackage{multirow}
\usepackage{mathtools}
\usepackage{cuted}
\usepackage{capt-of}
\usepackage[table,xcdraw]{xcolor}
\usepackage{algorithmic}
\usepackage[ruled,vlined]{algorithm2e}
\usepackage[normalem]{ulem}
\useunder{\uline}{\ul}{}
\usepackage{bbding}

\usepackage[pagebackref,breaklinks,colorlinks]{hyperref}

\def\eg{\emph{e.g}}
\def\etal{\emph{et al.}}

\def\ie{\emph{i.e.}}

\usepackage[capitalize]{cleveref}
\crefname{section}{Sec.}{Secs.}
\Crefname{section}{Section}{Sections}
\Crefname{table}{Table}{Tables}
\crefname{table}{Tab.}{Tabs.}

\def\cvprPaperID{6843} 
\def\confName{CVPR}
\def\confYear{2022}

\begin{document}

\title{Ray3D: ray-based 3D human pose estimation for monocular absolute 3D localization}

\author{Yu Zhan\\
Aibee Inc.\\

\and
Fenghai Li\\
Beijing Technology and Business University\\

\and
Renliang Weng\\
Aibee Inc.\\

\and
Wongun Choi\\
Aibee Inc.\\

}
\maketitle

\begin{abstract}
  In this paper, we propose a novel monocular ray-based 3D (Ray3D) absolute human pose estimation with calibrated camera. Accurate and generalizable absolute 3D human pose estimation from monocular 2D pose input is an ill-posed problem. To address this challenge, we convert the input from pixel space to 3D normalized rays. This conversion makes our approach robust to camera intrinsic parameter changes. To deal with the in-the-wild camera extrinsic parameter variations, Ray3D explicitly takes the camera extrinsic parameters as an input and jointly models the distribution between the 3D pose rays and camera extrinsic parameters. This novel network design is the key to the outstanding generalizability of Ray3D approach. To have a comprehensive understanding of how the camera intrinsic and extrinsic parameter variations affect the accuracy of absolute 3D key-point localization, we conduct in-depth systematic experiments on three single person 3D benchmarks as well as one synthetic benchmark. These experiments demonstrate that our method significantly outperforms existing state-of-the-art models. Our code and the synthetic dataset are available at \href{https://github.com/YxZhxn/Ray3D}{https://github.com/YxZhxn/Ray3D}.
\end{abstract}

\begin{figure}[htbp]
	\centering
	\includegraphics[width=0.47\textwidth]{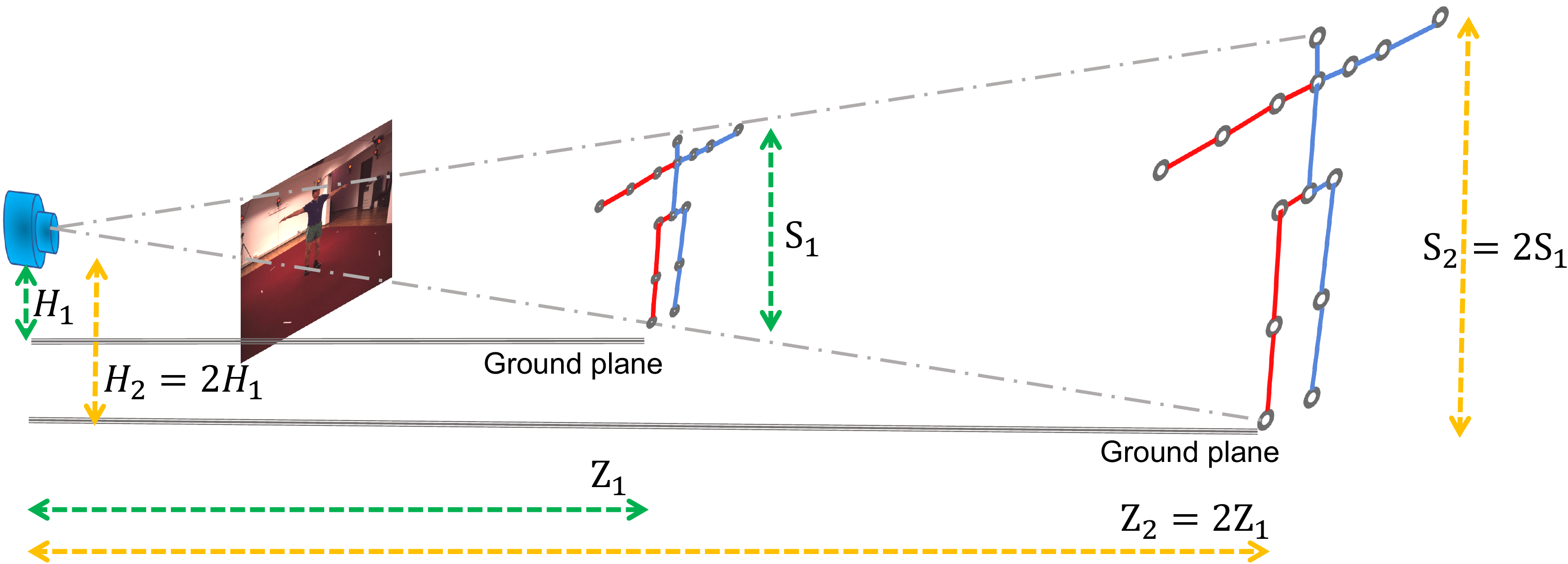}
	\begin{flushleft}
	    \vskip-12pt
		{\footnotesize \hskip100pt (a) } 
	\end{flushleft}
	\vskip-10pt
	\includegraphics[width=0.47\textwidth]{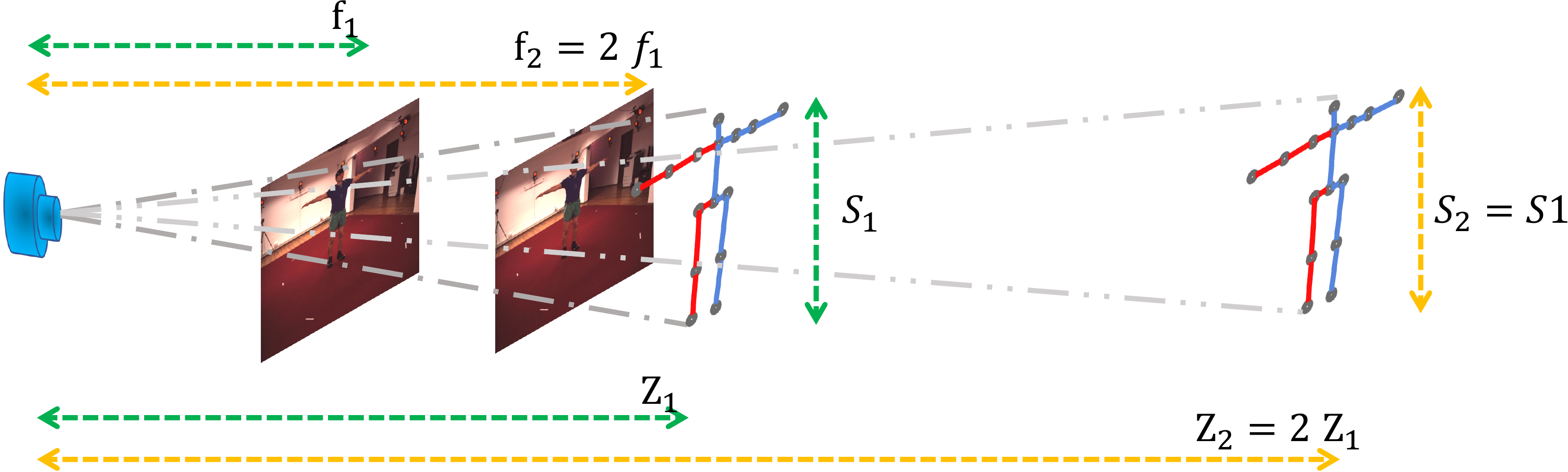}
	\begin{flushleft}
	    \vskip-12pt
		{\footnotesize  \hskip100pt (b) }
	\end{flushleft}
	\vskip-17pt
	\caption{As shown in (a), if both the body size and the distance to camera are scaled up by twice, the projected 2D keypoints locations remain the same. Same phenomenon is observed in (b), where both the focal length and 3D distance are doubled. ${Z_1}$ and ${Z_2}$ refer to the distance from the person to the camera, ${H_1}$ and ${H_2}$ represent the height of camera from the ground plane. ${S_1}$ and ${S_2}$ are the scale of the person. ${f_1}$ and ${f_2}$ represent focal length of the camera.} 
	\label{fig:ambiguity}
    \vspace{-6mm}
\end{figure}

\section{Introduction}
\label{sec:intro}
Accurate monocular 3D human pose estimation has found its wide applications in augmented reality~\cite{lin2010augmented}, human-object interaction~\cite{chen2019holistic}, and video action recognition~\cite{sijie2018spatial}. While the problem has been extensively studied in recent years, it's a well-known ill-posed problem~\cite{chen1985determination} with limited generalization capability. The problem becomes even more difficult when absolute 3D human pose estimation in a metric space is required, as knowing exactly where a human joint is in the World Coordinate System (WCS) is much more challenging than estimating the relative 3D offset of that joint from a reference point. While being more challenging, knowing absolute 3D poses is more desirable than the root-relative 3D poses in the real-world applications. For instance, unmanned store requires to detect the merchandise picked up by the customer, which relies on accurate hand localization in world coordinate system. 

A key-point's 2D pixel location is jointly determined by the scale of person's body figure, camera intrinsic parameters, camera extrinsic parameters and 3D position in the world coordinate system. These factors introduce ambiguities for 3D pose estimation. For instance, as shown in Figure~\ref{fig:ambiguity}~(a), if both the body size and the distance to camera are scaled up by twice, the projected 2D key-points locations remain the same. Similarly, if both the focal length and 3D distance are doubled, the 2D key-points keep the same, as illustrated in Figure~\ref{fig:ambiguity}~(b).
Typically, there are more than one configuration of 3D key-points that can generate the same observation of 2D key-points in the image plane.
Thus, naively learning a model to map from 2D pixel locations to 3D world locations is arguably prone to failure.  

To resolve these ambiguities, a number of monocular 3D human estimation approaches have been proposed~\cite{ju2019absposelifter,dario2019videopose,jianan2020smap,andrei2018deep,dabral2019multi,moon2019camera}. These methods can be mainly categorized into two groups, \ie, ~lifting methods and image based methods. Lifting methods ~\cite{yujun2019exploiting,ChengYWWT19,DabralMKASJ18,dario2019videopose,ZhangHW2020,ZhouH0XW17,LiuSW0CA20} take the 2D human poses as input and lift the 2D pose to 3D pose. A few lifting methods normalize the input according to image resolution~\cite{dario2019videopose}, and camera principal point~\cite{ju2019absposelifter}. While these normalization schemes improve the generalization ability to some extent, they fail to fully resolve the ambiguity due to variation in camera intrinsic parameters. On the other hand, image based approaches~\cite{moon2019camera,dabral2019multi,andrei2018monocular,lassner2017unite,smplify,zerong2019deephuman,nikos2019learning,aaron3d2018} estimate the 3D root position based on the prior about the body size. In contrast, \cite{jianan2020smap,moon2019camera} rely on image-based human depth estimation for absolute root-keypoint localization. The issue with these learning-based depth estimation approaches is lack of sufficient training data with viewpoint variations. For instance, the model trained with front-view viewpoint may not generalize well to cameras with large pitch value. Moreover, they fail to fully address the aforementioned ambiguities. 
	
To address the challenges more effectively, we propose our Ray3D method. Firstly, in order to have an intrinsic-parameter-invariant representation, we convert the 2D key-points in a pixel space to 3D rays in a normalized 3D space. With this simple design, our Ray3D approach achieves stable performance regardless of camera intrinsic parameter changes. 
Inspired by Videopose~\cite{dario2019videopose} and RIE~\cite{wenkang2021improving}, we fuse 3D rays from consecutive frames by using temporal convolution in order to further resolve the ambiguity introduced by occlusion and to improve accuracy. This temporal fusion mechanism stabilizes the output and generates more accurate 3D locations. Secondly, we jointly embed the camera extrinsic parameters into the network. Camera extrinsic parameters contain essential information for accurate 3D human pose estimation. Arguably, exploiting camera extrinsic parameters is the only way to resolve the human body part size ambiguity. For instance, in Fig.~\ref{fig:ambiguity} (a), if the camera's height is known to be close to $H_1$, we can safely eliminate incompatible hypotheses like the 3D pose with $S_2/H_2$. Therefore, it's essential to incorporate the camera extrinsic parameters into the network for accurate absolute localization. To our best knowledge, none of the existing learning-based 3D human pose estimation approaches explicitly utilize these information. In contrast, we directly take camera height and camera pitch value as input, and learn an independent camera embedding through a Multi-Layer Perceptron (MLP). This camera embedding is then concatenated with temporally fused ray features for 3D pose estimation. 

To understand and diagnose the absolute 3D pose estimators, we conduct a series of comprehensive and systematic experiments. Specifically, we explicitly benchmark the robustness of the approaches against focal length, principal point, camera pitch angle, camera height, camera yaw angle, body figure size variations on synthetic dataset. Furthermore, we evaluate generalization capability of these approaches on three single person benchmarks. 

To summarize, the proposed method makes the following contributions,
\begin{itemize}[noitemsep,nolistsep]
\item We convert the input space from 2D pixel space to 3D rays in a normalized coordinate system. This simple design effectively normalizes away the variations introduced by the camera intrinsic parameter changes as well as the camera pitch angle changes.
\item We present a novel and simple network which learns a camera embedding using the camera extrinsic parameters, and jointly models the distribution of camera extrinsic parameters and 3D rays. 
\item We provide a comprehensive and systematic benchmarking of existing 3D approaches in terms of robustness against camera pose variations, as well as cross-dataset generalization. 
\item Experiments on three real benchmark datasets and one synthetic dataset clearly demonstrates the advantages of our Ray3D approach.
\end{itemize}


\begin{figure*}[tbp]
    \centering
	\includegraphics[width=0.83\linewidth]{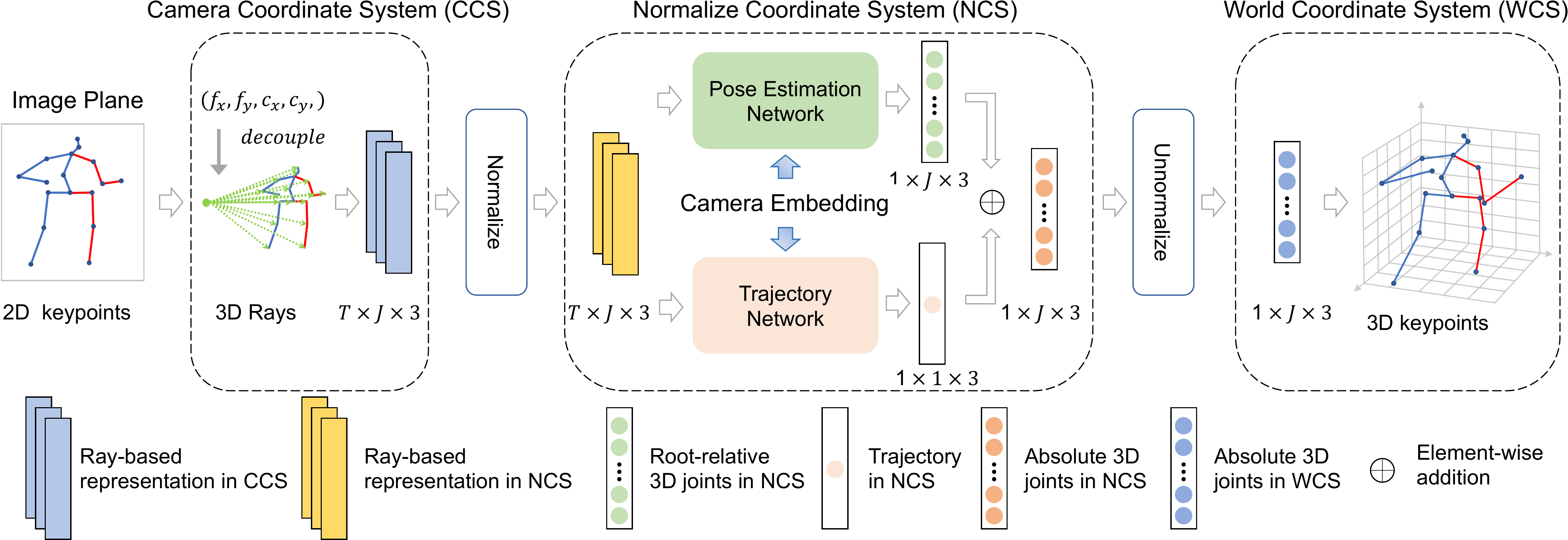} 
	\vskip-9pt
	\caption{An exposition of our Ray3D architecture. In pre-processing, we convert 2D input to ray-based 3D representation. These 3D rays are transformed to NCS, which are subsequently fed to pose estimation network and trajectory network to  predict the final absolute 3D pose. With unnormalization, the 3D pose under world coordinate system is obtained.}
	\label{fig:overview}
	\vspace{-6mm}
\end{figure*}

\section{Related Work}

\noindent\subsection{Lifting based 3D human pose estimation} 
Lifting based 3D human pose estimation approaches learn a model to map from 2D human pose to 3D. While only using 2D coordinates without image features may appear sub-optimal, these methods enjoy good performance-cost trade-off.

The majority of lifting based approaches work on root-relative 3D human pose estimation~\cite{martinez2017simple,ce2021poseformer,TomeRA17,wenkang2021improving,yujun2019exploiting,ChengYWWT19,DabralMKASJ18,dario2019videopose,ZhangHW2020,ZhouH0XW17,LiuSW0CA20}. ~\cite{martinez2017simple} is the pioneer work that introduces the lifting design. ~\cite{yujun2019exploiting,ce2021poseformer,dario2019videopose, wenkang2021improving} exploit temporal information to improve the 3D pose estimation accuracy, especially for the occluded cases. Pose ambiguities can be partially resolved by exploiting the temporal context. Shan~\etal ~\cite{wenkang2021improving} propose to encode relative positional and temporal enhanced representations, and this approach achieves excellent root-relative 3D pose accuracy. Inspired by ~\cite{wenkang2021improving}, we encode relative 3D normalized rays to improve the root-relative pose model.

Only a handful of lifting approaches can be applied for absolute 3D human pose estimation. Pavllo~\etal~\cite{dario2019videopose} employ a trajectory model to estimate the 3D trajectory of the root joint. Chang~\etal ~\cite{ju2019absposelifter} normalize the input by subtracting principal point of the camera and reconstruct up to the canonical root depth. This root depth is further multiplied with focal length to generate the final absolute depth. While these approaches achieve promising results, they fail to fully normalize the input by using camera intrinsics. Meanwhile, camera extrinsics are simply ignored. Thus, their performance is susceptible to camera intrinsics/extrinsics variations.  

\noindent\subsection{Image based 3D human pose estimation} 
Image based approaches aim to improve the 3D estimation accuracy by directly utilizing image features~\cite{ijcnn2019abs,wang2020hmor,jianan2020smap,icann2020multi,vibe,BCNet2020,zerong2019deephuman,nikos2019learning,zhang2021pymaf,learning2019thiemo,KolotourosPD19}. Rogez~\etal~\cite{lcr2017cvpr} frame the estimation problem as pose proposal generation, proposal scoring and proposal refinement. The 3D location of the root joint is obtained by minimizing the distance between 2D pose and projected 3D pose. Moon ~\etal~\cite{moon2019camera} devise a network to estimate the depth of root joint from cropped single person image, which inevitably loses contextual information of the subject. Alternatively, ~\cite{ijcnn2019abs,wang2020hmor,jianan2020smap,icann2020multi} estimate the root depth from the full image up to a scale. The issue with this direction is the requirement of significant amount of training data with camera variations to make image-based depth estimation reliable. Meanwhile, camera extrinsics are not taken into account, which limits their generalizability. Additionally,~\cite{Kocabas21} firstly proposes to estimate perspective camera for 3D body pose regression. ~\cite{Kocabas21} tries to estimate camera parameters along with the pose as well, which improves the generalizability.

 
\noindent\subsection{Camera encoding} 
Few approaches have explicitly encoded camera extrinsics to assist the vision tasks. Nerf~\cite{nerf,pixelnerf2021yu,mvsnerf} is a popular 3D object reconstruction approach that directly takes 2D camera viewing angle into input. The 2D viewing angle (\ie, pitch and yaw) is concatenated with 3D location of object point and then processed by a multilayer perceptron network. Differently, our approach learns an camera embedding specifically for camera pitch and camera height from the ground plane. This embedding largely resolves the ambiguities in the absolute 3D human pose estimation problem. 

\section{Proposed Method}

Intuitively, accurate monocular 3D absolute pose estimation relies on sufficient ambiguity reduction. Our method is designed to resolve ambiguities introduced by camera intrinsic parameter variation, body occlusion and camera pose variations with normalized representation of keypoints, temporal convolution and camera embedding correspondingly. 

In Figure~\ref{fig:overview}, we present the overview of the proposed Ray3D framework. To eliminate the impact of the intrinsic parameter variation, the 2D key-points are converted into 3D rays in Camera Coordinate System (CCS). To deal with the camera pitch angle variation, we further transform these 3D rays into (pitch) \emph{Normalized} Coordinate System (NCS). Similarly, ground truth 3D poses are transformed to NCS as well. In this way, both the input and output of the model are aligned into the same coordinate system. 

Temporal key-point motion information helps resolve 3D pose estimation ambiguity introduced by the occlusion~\cite{dario2019videopose,wenkang2021improving}. Following ~\cite{wenkang2021improving}, we fuse 3D rays from consecutive frames temporally and encode the relative pose rays to capture motion information. Different from ~\cite{wenkang2021improving}, our Ray3D approach learns a camera embedding which can (implicitly) provide strong constraints to eliminate ambiguities in absolute 3D pose estimation. Specifically, we employ an MLP network to learn a compact embedding for camera pose representation. This camera embedding is subsequently concatenated with latent 3D ray features for the pose prediction. This novel design largely improves the model's robustness against camera pose and body scale variations.

Following ~\cite{dario2019videopose}, we decompose the problem into two sub-problems, ~\ie, root-relative pose estimation and root location estimation. These two sub-problems are solved by our pose network and trajectory network respectively. Specifically, the relative 3D pose generated by pose estimation network is added to the root joint coordinate predicted by trajectory network. Finally, the human pose are \emph{unnormalized} into World Coordinate System (WCS).

\begin{figure}[tbp]
    \centering
	\includegraphics[width=0.8\linewidth]{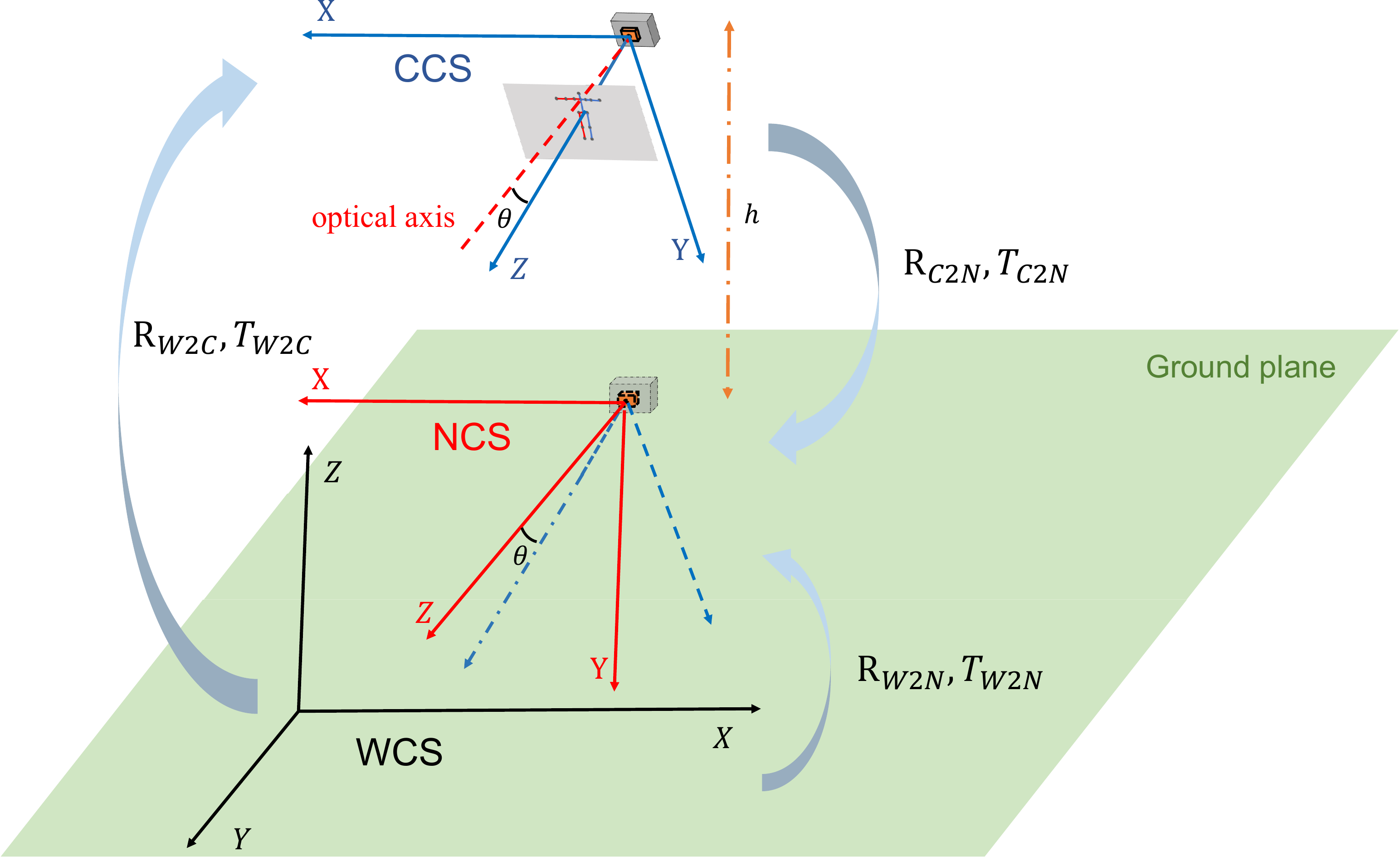}
	\vskip-9pt
	\caption{Normalized camera coordinate system is acquired by rotating the camera coordinate system along the \emph{x} axis with degree of ${\theta}$ and translating along the \emph{z} axis of the world coordinate system with distance of \emph{h}. \emph{h} is the height of camera in WCS. Such that the input and output of the lifting network are aligned in the same coordinate system.}
	\label{fig:normalized}
	\vspace{-4mm}
\end{figure}

\noindent\subsection{Input pre-processing}
\label{sec:preprocessing}

\subsubsection*{Intrinsic parameter decoupling}
\label{subsec: intrinsic}

Lifting based 3D pose estimation approaches manage to lift predicted 2D key-points ${\{p_i\}^J_{i=1}}$ to 3D key-points ${\{P_i^C\}^J_{i=1}}$ with deep neural network. $p_i=[x_i, y_i]$ stands for the location of the $i$ th joint of the person in the input image coordinate system and $P_i^C=[X_i^C, Y_i^C, Z_i^C]$ represents the corresponding joint in CCS. $J$ denotes the indices of joints. In order to achieve the invariance to the  camera intrinsic parameter changes, we perform the following transformation to ${\{p_i\}^J_{i=1}}$ (camera un-distortion can be added if needed):
\begin{equation}
    x^{ray}_{i} = \frac{x_i - c_x}{f_x}, y^{ray}_{i} = \frac{y_i - c_y}{f_y}, z^{ray}_{i} = 1.
    \label{eq:decouple}
\end{equation}
Such that we have 3D rays $\{p^{ray}_i\}^J_{i=1} = \{[x^{ray}_{i}, y^{ray}_{i}, z^{ray}_{i}]\}^J_{i=1}$. In Eq.~\ref{eq:decouple}, $c_x$ and $c_y$ represent for camera center points, $f_x$ and $f_y$ denote the focal length. ${p^{ray}_i}$ is a \emph{ray} that points from optical center of the camera to the key-point ${i}$ in the image plane. 

Unlike \cite{ju2019absposelifter}, we completely eliminate the impact of focal length by explicitly normalize the ray representation with the calibrated focal length. Compared to~\cite{Cho_2021_ICCV}, our 3D rays are converted to normalized rays in Normalized Coordinate System (NCS), which will be shortly discussed in the next subsection.

\subsubsection*{Extrinsic parameter decoupling with the Normalized Coordinate System (NCS)} 
\label{subsec: normalization}

Define a key-point in the world, camera, and \emph{normalized} coordinate system as ${P_W}$, ${P_C}$ and ${P_N}$ respectively. With accurate calibration, one can acquire camera extrinsics including rotation matrix ${R_{W2C}}$, and translation vector ${T_{W2C}}$. And the transformation between ${P_W}$ and ${P_C}$ is as follows:
\begin{equation}
    P_C = R_{W2C}\cdot P_W + T_{W2C}.
    \label{eq:w2c}
\end{equation}

\begin{figure}[htbp]
    \centering
	\includegraphics[width=1\linewidth]{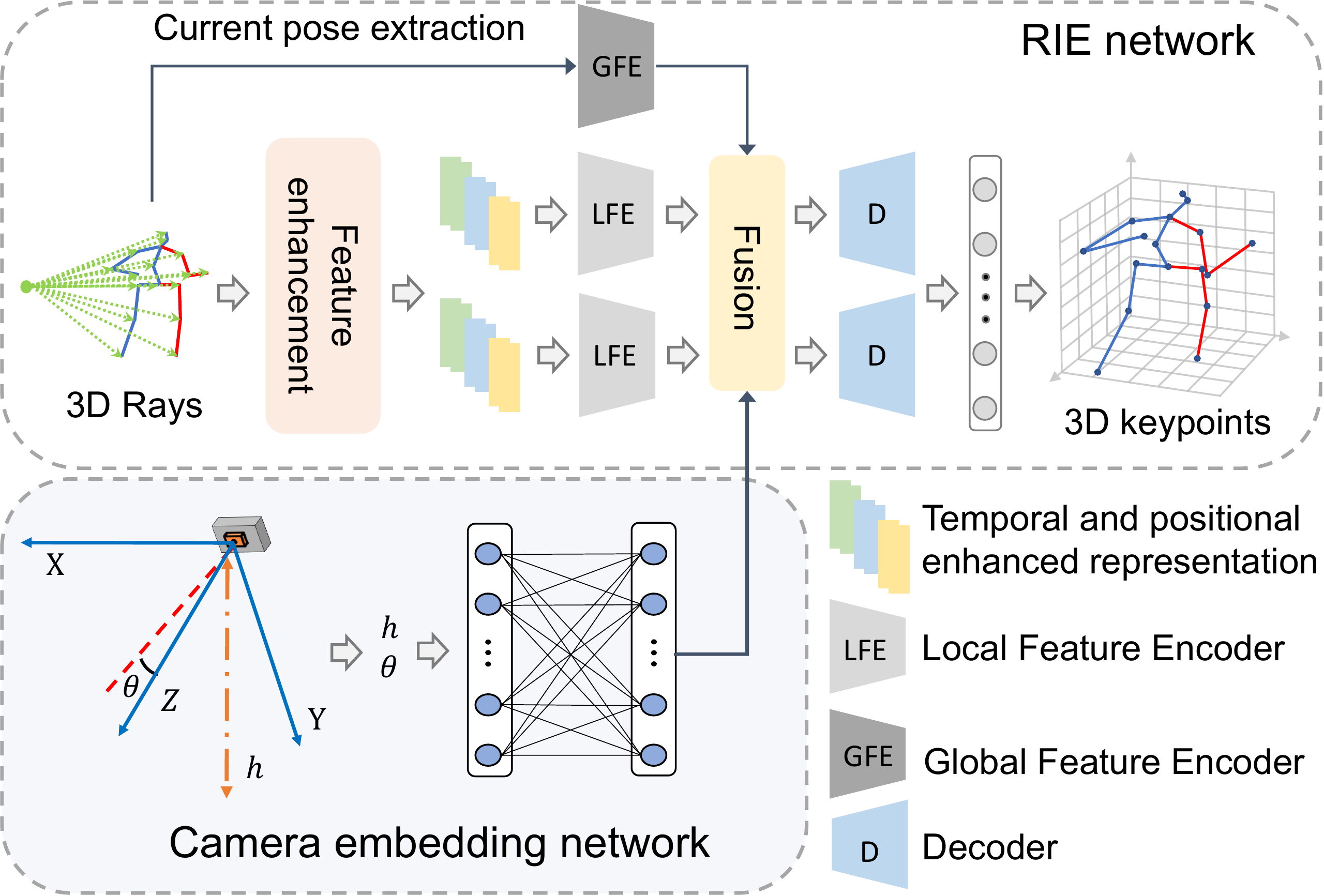} 
	\vskip-9pt
	\caption{Overview of our lifting network. Our relative pose estimation and root joint estimation networks share the same RIE architectures. The network is equipped with positional and temporal enhanced representation. For more details, please refer to~\cite{wenkang2021improving}. MLP based camera embedding works as a plug-in to generate embedding features, and then concatenate with latent ray features for final pose prediction.}
	\label{fig:embedding}
	\vspace{-3mm}
\end{figure}


In this paper, we aim to predict absolute 3D human pose in WCS. Camera pose in the 3D world can be determined by its 3D location, pitch, yaw and roll angles. Pitch, ${\theta}$, describes the angle between the optical axis of the camera and the ground plane. With an assumption that camera yaw and roll values are close to 0, 
the camera pitch value and the camera height can uniquely specify the pose of a camera up to horizontal translation. In order to explicitly encode pitch for accurate pose estimation, we set up NCS as presented in Fig.~\ref{fig:normalized}. First, the CCS is rotated along the ${x}$ axis of CCS to eliminate the pitch angle. Then, the coordinate system is translated along the ${y}$ axis of CCS to the ground plane.  

One can easily calculate the rotation matrix and translation vector between ${P_C}$ and ${P_N}$:
\begin{equation}
R_{C2N} = 
\left(
    \begin{matrix}
        1 & 0 & 0  \\
        0 & \cos{\theta} & \sin{\theta}  \\
        0 & -\sin{\theta} & \cos{\theta}  \\
    \end{matrix}
\right),
\label{eq:rc2n}
\end{equation}
\begin{equation}
T_{C2N}  = 
\left(
    \begin{array}{ccc}
        0 & -h & 0 \\
    \end{array}
\right).
    \label{eq:tc2n}
\end{equation}
According to the Eq.~\ref{eq:w2c},~\ref{eq:rc2n} and~\ref{eq:tc2n}, we have:
\begin{equation}
    R_{W2N} = R_{C2N} \cdot R_{W2C} ,
    \label{eq:rw2n}
\end{equation}
\begin{equation}
    T_{W2N} = R_{C2N} \cdot T_{W2C} +T_{C2N} ,
    \label{eq:tw2n}
\end{equation}
\begin{equation}
    P_N = R_{C2N} \cdot P_C +T_{C2N} ,
    \label{eq:c2n}
\end{equation}
\begin{equation}
    P_N = R_{W2N}\cdot P_W + T_{W2N}.
    \label{eq:w2n}
\end{equation}
By applying the Eq.~\ref{eq:c2n} to ${\{p_i^{ray}\}^J_{i=1}}$ and Eq.~\ref{eq:w2n} to ground-truth 3D keypoints ${\{P_i^W\}^J_{i=1}}$, we can get normalized 3D rays ${\{\hat{p}_i^{ray}\}^J_{i=1}}$ and normalized 3D ground-truth ${\{P_i^N\}^J_{i=1}}$. As a result, our Ray3D network is trained to lift from  ${\{\hat{p}_i^{ray}\}^J_{i=1}}$ to ${\{P_i^N\}^J_{i=1}}$ within the same coordinate system, which reduces the training difficulty and increases model robustness.

\subsection{Lifting network}
\label{sec:network}



\subsubsection*{Absolute pose estimation} 
\label{subsec: abs_pose}
The task of estimating 3D absolute human poses is composed of two sub-problems, root location estimation (i.e., estimating the location of the center of mass of a person) and root-relative pose estimation (i.e., offset of each key point with respect to the center). We design our network to jointly learn to solve the two sub-problems with the trajectory network and the pose network, respectively (see Fig.~\ref{fig:overview}). The outputs of these two networks are added to produce the absolute 3D poses.

Temporal motion information improves model's robustness against body occlusion. Inspired by \cite{wenkang2021improving}, we adopt RIE architecture as our backbone network for both the root-relative pose network and trajectory network. As shown in Fig.~\ref{fig:embedding}, RIE network is enhanced with positional and temporal information. Relative positions of input key-points are encoded as positional information within the frame, and differences between the 2D pose of current frame and the one from neighbouring frames are treated as temporal information. Such enhanced input are divided into $5$ groups (torso, left arm, right arm, left leg, and right leg) for local feature learning. In addition, global feature is extracted from current frame to maintain the consistency of overall posture. Feature fusion module aggregates all these features for 3D pose estimation. Using this architecture, we replace vanilla 2D human key-points with our intrinsic-invariant normalized 3D rays as input for both pose network and trajectory network to address ambiguities. We refer the reader to ~\cite{wenkang2021improving} for the details of RIE structure. Note that the contribution of this work is not on the specific network design, rather on the input representation and explicit embedding of camera extrinsic parameters. This novel design can be easily incorporated to the existing pose estimation framework. 

\subsubsection*{Camera embedding} 
\label{subsec: embedding}
We argue that camera extrinsic parameters are critical for absolute pose estimation in WCS, and propose to explicitly utilize extrinsic parameters by learning an independent camera embedding through a Multi-Layer Perceptron with $\theta$ and $h$ as inputs. Specifically, the camera embedding module is constructed with two fully connected layers followed with batch normalization~\cite{IoffeS15}, rectified linear unit~\cite{NairH10} and Dropout~\cite{SrivastavaHKSS14}. As shown in Fig.~\ref{fig:embedding}, this camera embedding is concatenated with temporally fused latent 3D ray features in both relative pose prediction and trajectory network. Therefore, both networks exploit camera extrinsic parameters for robust and accurate pose estimation.

\begin{table*}[htbp]
\centering
\tiny
\caption{Quantitative evaluation results under MPJPE on H36M using GT keypoints as input. (f = 9) means this approach utilizes 9 consecutive frames for pose estimation, and (f = 1) means the approach does not make use of temporal information. Best results are shown in \textbf{bold}.}
\vskip-12pt
\begin{tabular}{@{}l|llllllllllllllll|l@{}}
\toprule
MPJPE          &                 & Dir.   & Disc. & Eat.  & Greet & Phone & Photo & Pose  & Purch. & Sit   & SitD. & Somke & Wait  & WalkD.& Walk  & WalkT. & Average\\ \midrule
Hossain et al. \cite{hossain2018exploiting}       &ECCV’18 & 35.2  & 40.8  & 37.2  & 37.4  & 43.2  & 44.0  & 38.9 & 35.6    & 42.3  & 44.6  & 39.7  & 39.7  & 40.2  & 32.8  & 35.5   & 39.2   \\
Liu et al. (f = 243) \cite{LiuSW0CA20}.     &CVPR’20 & 34.5  & 37.1  & 33.6  & 34.2  & 32.9  & 37.1  & 39.6 & 35.8    & 40.7  & 41.4  & 33.0  & 33.8  & 33.0  & 26.6  & 26.9   & 34.7   \\
Videopose. (f = 9) \cite{dario2019videopose}   &CVPR'19 & 37.0  & 40.7  & 35.2  & 37.4  & 38.4  & 44.2  & 42.3 & 37.1    & 46.5  & 48.8  & 38.9  & 40.1  & 38.5  & 29.9  & 32.6   & 39.2   \\
PoseFormer (f = 9) \cite{ce2021poseformer}        &ICCV'21 & 49.2  & 49.7  & 38.7  & 42.7  & 40.0  & 40.9  & 50.7 & 42.2    & 47.0  & 46.1  & 43.4  & 46.7  & 39.8  & 36.4  & 38.0   & 43.5   \\
PoseAug (f = 1) \cite{GongZF21}                   &CVPR'21 & -     & -     & -     & -     & -     & -     & -    & -       & -     & -     & -     & -     & -     & -     & -      & 38.2   \\
RIE (f = 9)~\cite{wenkang2021improving}                   &ACMMM'21 & 34.8  & 38.2  &\textbf{31.1}  & 34.4  & 35.4  & \textbf{37.2}  & 38.3  & 32.8  & \textbf{39.5} & \textbf{41.3} & 34.9   & \textbf{35.6}  & 32.9  & \textbf{27.1}  & \textbf{28.0}   & 34.8   \\ \hline
Ray3D (f = 9)                   &                 & \textbf{31.2}  & \textbf{35.7}  & 31.4 & \textbf{33.6} & \textbf{35.0}  & 37.5  & \textbf{37.2} & \textbf{30.9} & 42.5  & \textbf{41.3} & \textbf{34.6}   & 36.5  & \textbf{32.0}   & 27.7  & 28.9  & \textbf{34.4}   \\ \bottomrule
\end{tabular}
\label{table:1}
\\

\end{table*}

\section{Experiments and results}
The evaluation results of the proposed method with different experiment setups are reported in this section. First, datasets and evaluation metrics are introduced in Section~\ref{subsec: exp_4_1}, and details of implementation is described in Section~\ref{subsec: exp_4_2}. Section~\ref{subsec: exp_4_3} showcases the comparison of our Ray3D and other state-of-the-arts on three public benchmarks. Then, generalization test result on a synthetic dataset is described in Section~\ref{subsec: exp_4_4}. Furthermore, the effectiveness of Ray3D's components is analyzed with ablation study in Section~\ref{subsec: exp_4_5}. Finally, the limitation of Ray3D is discussed in Section~\ref{exp: discuss}.
\subsection{Datasets and Evaluation metrics}
\label{subsec: exp_4_1}

We evaluate our Ray3D on three public datasets captured with different camera poses and human poses. The camera intrinsics and extrinsics are provided by all datasets. \emph{The following datasets contain personal identifiable information about human subjects. All the subjects in these datasets have granted their permission for the dataset creation.}

\noindent\textbf{Human3.6M (H36M)}~\cite{ionescu2013human3} is a large-scale 3D human pose estimation dataset, which contains 3.6 million video frames recorded with four synchronized cameras. Following previous works \cite{dario2019videopose,ce2021poseformer}, five subjects (S1, S5, S6, S7, S8) and two subjects (S9, S11) with 17-key-point definition are used as training and testing data respectively for SOTA comparison.

\noindent\textbf{Humaneva-I}~\cite{SigalBB10} is a much smaller dataset compared with H36M , which is captured in a controlled indoor environment with three cameras. The proposed Ray3D representation requires well-calibrated intrinsic and  extrinsic, as a result, Camera 2 and Camera 3 are removed due to bad camera calibration.

\noindent\textbf{MPI-INF-3DHP (3DHP)}~\cite{mehta2017monocular_3dhp} consists of 1.3 million video frames, which covers more diverse human motions than Human3.6M. Following previous work~\cite{GongZF21}, poses with 17 joints from Camera 0, 1, 2, 4, 5, 6, 7 and 8 are used for training. Sequence of TS1, TS3 and TS4 are adopted as test sets. TS2, TS5 and TS6 are excluded due to inaccurate or incomplete camera calibration.

In our experiments, we adopt following evaluation metrics: Mean Per Joint Position Error (MPJPE) in millimeters is used to evaluate root relative pose estimation results under CCS. To evaluate the performance of absolute pose under WCS, Absolute MPJPE (Abs-MPJPE) is adopted which calculates the difference between the prediction and GT pose in WCS. Mean of the Root Position Error (MRPE) proposed by~\cite{ju2019absposelifter} is used to evaluate the root joint's trajectory prediction.

\begin{table*}[htbp]
\centering
\tiny
\caption{Quantitative evaluation results under Abs-MPJPE and MRPE on H36M using CPN detected keypoints as 2D input. Best results are shown in \textbf{bold}.}
\vskip-12pt
\begin{tabular}{@{}l|llllllllllllllll|l@{}}
\toprule
Abs-MPJPE              &                                   &Dir.  &Disc. &Eat.  &Greet &Phone &Photo &Pose   &Purch   &Sit   &SitD.  &Somke &Wait  &WalkD. &Walk  &WalkT. &Average \\ \midrule
Videopose (f = 9) ~\cite{dario2019videopose} &CVPR'19 &128.9 &125.4 &124.4 &138.2 &\textbf{108.2} &155.5 &116.6  &101.1   &135.8 &287.6  &128.6 &130.9 &122.1  &101.6 &110.7  &134.4  \\
PoseLifter (f = 1)~\cite{ju2019absposelifter}           &ICCV’19 &140.9 &113.2 &139.9 &148.2&122.0  &155.3 &121.5  &121.1   &170.0 &267.6  &139.2 &142.9 &146.4  &132.1 &135.2  &146.4   \\
PoseFormer (f = 9) \cite{ce2021poseformer}        &ICCV'21 &112.6 &137.1 &\textbf{117.6} &145.8 &113.0 &166.0 &125.5  &113.8   &\textbf{128.8} &245.7  &122.7 &144.8 &125.0  &118.9 &129.3  &136.5  \\
RIE (f = 9) ~\cite{wenkang2021improving}          &ACMMM'21 &143.2 &133.2 &143.9 &142.7 &110.9 &151.4 &125.9  &98.4    &136.4 &273.4  &127.5 &138.9 &126.8  &107.3 &116.0  &138.4   \\ \hline
Ray3D (f = 1)                                     &        &\textbf{80.1}  &100.8 &123.8 &125.5 &110.7 &111.8 &96.1   &99.3    &129.4 &176.3  &\textbf{106.8} &129.2 &120.4  &109.1 &106.6. &115.1 \\
Ray3D (f = 9)                                     &        &92.9  &\textbf{97.4}  &139.8 &\textbf{118.6} &113.8 &\textbf{105.9} &\textbf{84.5}   &\textbf{74.9}   &148.6 &\textbf{165.7}  &116.6 &\textbf{113.9} &\textbf{98.2}   &\textbf{83.6}  &\textbf{87.9}  &\textbf{109.5} \\ \bottomrule
MRPE                    &                                  &Dir.  &Disc. &Eat.   &Greet &Phone &Photo &Pose  &Purch.  &Sit   &SitD.  &Somke &Wait  &WalkD. &Walk  &WalkT. &Average \\ \midrule
Videopose. (f = 9) ~\cite{dario2019videopose} &CVPR'19  &124.2 &115.9 &111.0 &127.3  &97.6  &141.9 &105.7 &96.4    &122.0 &276.5  &119.6 &123.3 &111.3  &94.0  &101.6  &124.6   \\
PoseLifter (f = 1)~\cite{ju2019absposelifter}.          &ICCV’19  &134.7 &102.3 &126.9 &135.7  &109.9 &138.5 &110.7 &110.9   &170.0 &252.4  &128.4 &133.9 &139.4  &121.6 &124.4  &135.1   \\
PoseFormer (f = 9) \cite{ce2021poseformer}       &ICCV'21  &104.7 &134.7 &\textbf{103.9} &137.4  &99.6  &154.6 &119.8 &108.9   &\textbf{108.2} &233.7  &111.1 &141.1 &116.2  &117.9 &123.8  &127.7   \\
RIE (f = 9) ~\cite{wenkang2021improving}         &ACMMM'21  &139.1 &124.5 &129.9 &133.1  &\textbf{99.2}  &141.4 &116.3 &93.5    &124.0 &265.9  &118.4 &131.3 & 117.1 &100.4 &109.2  &129.6  \\ \hline
Ray3D (f = 1)                                    &         &\textbf{67.3}  &91.7  &113.6 &111.8  &104.5 &96.3  &85.8  &94.6    &124.4 &161.7  &\textbf{97.6}  &119.5 &110.9  &100.9 &94.8   &105.0\\
Ray3D (f = 9)                                    &         &83.7  &\textbf{86.8}  &128.9 &\textbf{104.8}  &109.3 &\textbf{91.6}  &\textbf{75.0}  &\textbf{65.2}    &143.9 &\textbf{150.5}  &108.6 &\textbf{105.7} &\textbf{88.4}  &\textbf{73.9}  &\textbf{77.8}  &\textbf{99.6} \\ \bottomrule
\end{tabular}
\label{table:2}

\end{table*}

\subsection{Implementation Details}
\label{subsec: exp_4_2}
For our Ray3D approach, dimension of camera embedding is set as 64. Initial learning rate is $0.001$. Adam optimizer with exponential learning rate decay factor of $0.99$ is employed. Horizontal flip augmentation is adopted both in training and testing. For H36M dataset, we adopt Cascaded Pyramid Network (CPN) ~\cite{cpn} detected poses and GT 2D poses as input. As for Humaneva-I and 3DHP, only GT 2D poses are used.

\subsection{Evaluation on public benchmarks}
\label{subsec: exp_4_3}
In this section, we first compare our Ray3D with state-of-the-art methods under all 15 action sequences on H36M, then generalizability of these methods is evaluated by cross-dataset testing. The comparing approaches include the latest PoseFormer~\cite{ce2021poseformer}, Videopose~\cite{dario2019videopose}, PoseLifter~\cite{ju2019absposelifter} and RIE~\cite{wenkang2021improving}. Note that PoseLifter was designed for absolute pose estimation. Different from Videopose and PoseLifter, both PoseFormer and RIE are incapable for absolute pose estimation. To test their capability for root-joint localization, we equip them with a trajectory model using their own network structure. For fair comparison, we carefully re-trained PoseFormer, Videopose, PoseLifter and RIE with their provided source code under PyTorch~\cite{NEURIPS2019_9015}. 

\noindent\textbf{H36M evaluation}
Table~\ref{table:1} shows the performance of the methods that focus on root-relative pose prediction where ground truth 2D keypoints are taken as input. From the table, we can observe that our Ray3D obtains comparable results compared to SOTA methods. Specifically, MPJPE surpasses RIE~\cite{wenkang2021improving} by 0.4mm. Table~\ref{table:2} shows the results for absolute pose estimation in WCS using CPN~\cite{cpn} detected 2D poses on H36M dataset. It can be seen that Ray3D outperforms all SOTA methods for both Abs-MPJPE and MRPE with clear margin. Compared with RIE, our method reduces Abs-MPJPE by 28.9mm and MRPE by 30.0mm respectively. It is worth noting that Ray3D outperforms PoseLifter by 31.3mm under Abs-MPJPE when no temporal information is used. These results demonstrate that Ray3D is effective and generates more accurate absolute 3D locations. 

Among these four baseline methods, PoseLifter using single frame performs the worst. And Ray3D working with 9 frames surpasses Ray3D using single frame. This verifies the benefits of using temporal features for 3D pose estimation. Another interesting observation is that these baseline methods perform similarly in terms of MRPE. This shows the network structure design is not the critical factor for absolute pose estimation. Rather the input representation and camera embedding are the keys for accurate keypoint localization. 

\begin{table*}[htbp]
\caption{Cross dataset evaluation. We adopt a 14-joint skeleton training on 3DHP, testing on H36M, Humaneva-I and 3DHP datasets.
MPJPE, Abs-MPJPE and MRPE are reported. The unit of all numbers is mm. The best results are in \textbf{bold}. }
\tiny
\centering
\small
\vskip-12pt
\begin{tabular}{@{}l|lcc|lll|lcc@{}}
\toprule
method \textbackslash datasets & \multicolumn{3}{c|}{H36M} & \multicolumn{3}{c|}{HumanEva-I} & \multicolumn{3}{c}{3DHP} \\ 
                               & MPJPE  & Abs-MPJPE   & MRPE    & MPJPE   & Abs-MPJPE  & MRPE  & MPJPE          & Abs-MPJPE  & MRPE  \\ \midrule
Videopose (f = 9)~\cite{dario2019videopose}                      & \textbf{81.2}   & 1680.3      & 1686.6  & 86.2    & 1387.4     & 1387.1& 58.2           & 149.1      & 143.0 \\
PoseFormer (f = 9)~\cite{ce2021poseformer}                     & 97.8   & 1824.0      & 1818.9  & 104.7   & 1470.4     & 1452.0& 47.3           & 207.5      & 211.3 \\
PoseLifter (f = 1)~\cite{ju2019absposelifter}                    & 92.9   & 573.3       & 570.5   & 240.2   & 1263.0     & 1129.3& 76.7           & 147.8      & 133.6 \\
RIE (f = 9)~\cite{wenkang2021improving}                           & 91.2   & 1679.9      & 1673.0  & 92.0    & 1375.8     & 1369.5& 50.9           & 135.6      & 132.4 \\
CDG (f = 1) ~\cite{WangSF20} & 95.6 & - & - & - & - & - & 90.3 & - & - \\
Ray3D (f = 9)                         & 84.4   & \textbf{243.9}       & \textbf{246.7}   & \textbf{83.9}    & \textbf{477.8}      & \textbf{468.6} & \textbf{46.6}           & \textbf{103.3}      & \textbf{95.3}  \\ \bottomrule
\end{tabular}
\label{table:cross_3dhp}
\end{table*}

\noindent\textbf{Cross-dataset testing}
We train comparing models in 3DHP dataset, and evaluate them using H36M and Humaneva-I. 14-joint definition is applied for all datasets during cross-dataset testing. For H36M and 3DHP, we remove mid spine, neck and chin keypoints. As for Humaneva-I, the thorax key-point is removed out of original 15 joints. As shown in Table~\ref{table:cross_3dhp}, none of the baselines work well in cross-scenario situations while the Ray3D shows good generalization performance in Humaneva-I and 3DHP dataset. This is because the camera intrinsic and extrinsic vary greatly across different scenes. Our Ray3D approach explicitly takes extrinsics(\ie, camera pitch and camera height) as input to learn a camera embedding, which results in improved generalization ability.

\subsection{Evaluation on synthetic dataset}
In this section, we conduct in-depth systematic experiments to benchmark 3D pose estimators' robustness against camera intrinsic, camera rotation (yaw), camera pitch, camera translation and person scale variations on a carefully curated synthetic benchmark. We use cameras from H36M to conduct camera augmentations. After the synthesis, the focal length of simulated cameras ranges from $1100$ to $1180$, where focal length in the training data ranges from $1143$ to $1150$.  Camera rotation ranges from $0$ to $360$ degrees. Camera pitch ranges from $0$ to $40$ degrees. Camera translation ranges from $9$ to $14$ meters. The total length of human limbs ranges from $2.5$ to $4.5$ meters (roughly, the height of human ranges from $1$ to $2$ meters). Specifically, $100$ virtual cameras are generated with fixed extrinsic for intrinsic generalization test, and $126$ virtual cameras are generated with fixed intrinsic for extrinsic generalization test. We additionally simulated $324$ cameras for training as well. Note that training and testing camera poses do not overlap. Five subjects (S1, S5, S6, S7, S8) and two subjects (S9, S11) with 14-key-point-definition are used for training and testing respectively. The detailed camera augmentation setting for training and testing are listed in the supplementary materials. To evaluate the effectiveness of camera embedding, we add a new baseline named Ray3D\_w/o\_CE where camera embedding branch is removed from Ray3D.
\label{subsec: exp_4_4}

\begin{figure}
  \centering
  \includegraphics[width=0.48\linewidth]{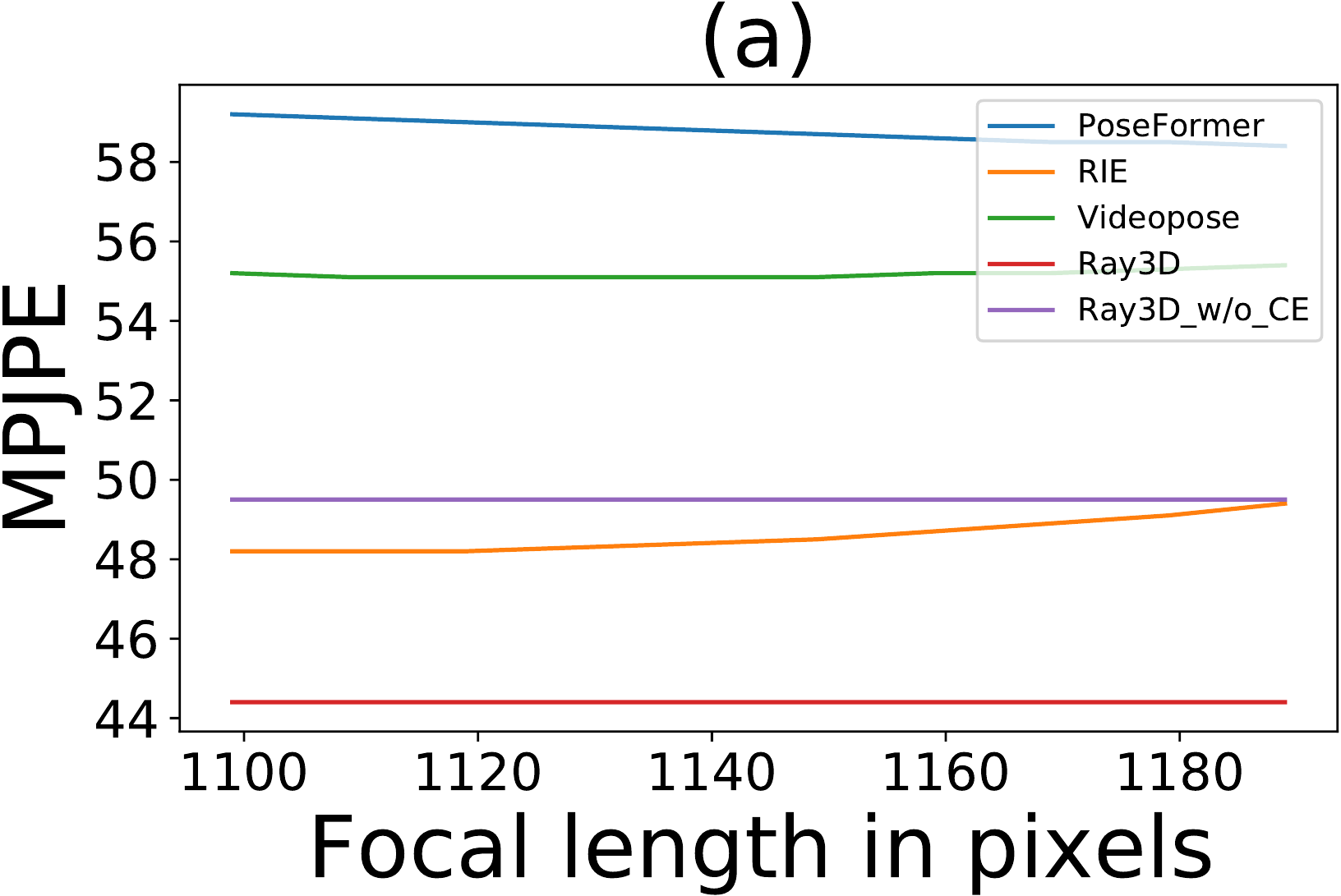}
  \includegraphics[width=0.48\linewidth]{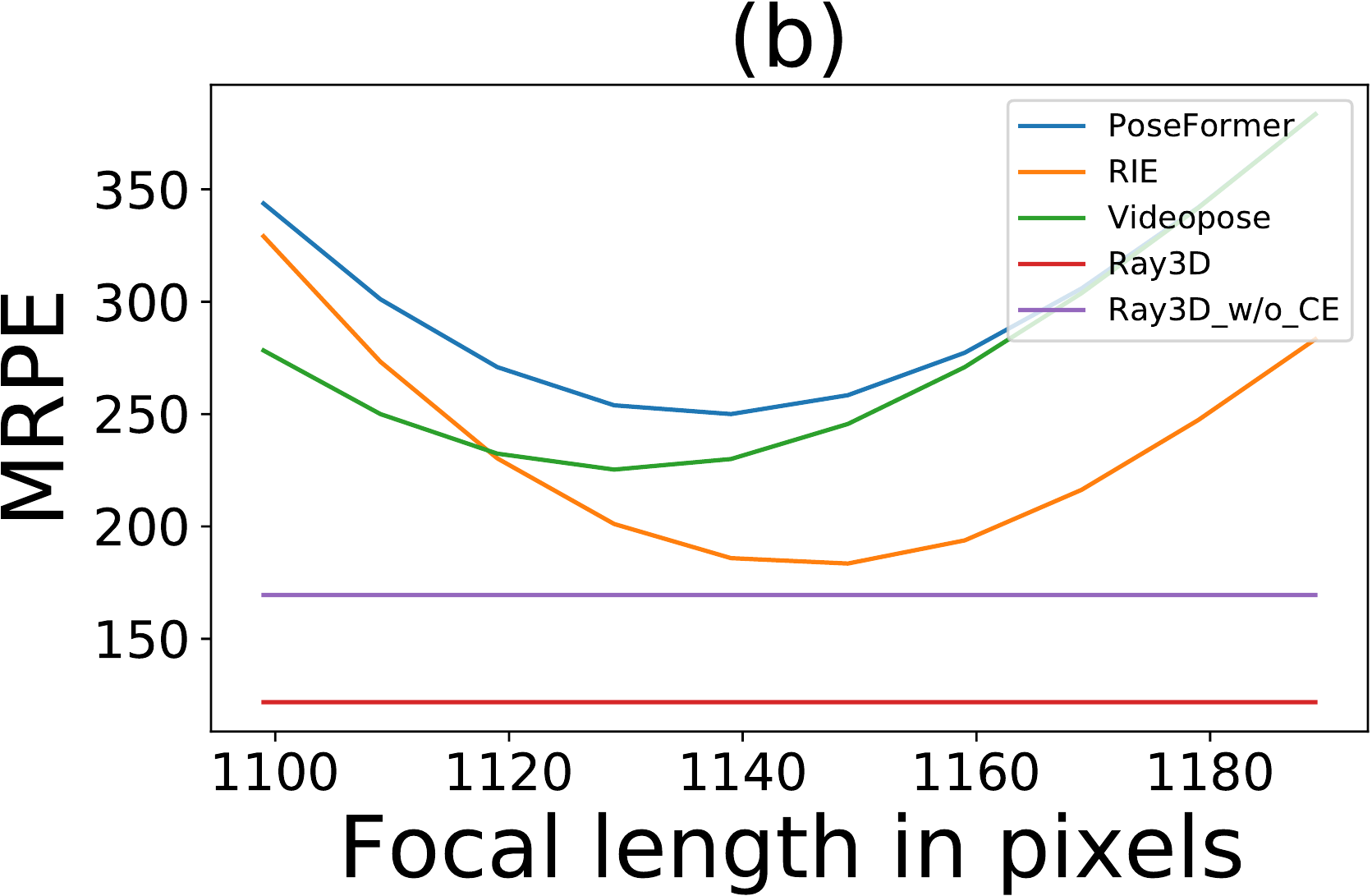}
  \vskip-9pt
  \caption{performance under MPJPE and MRPE in case of focal length changes are plotted in (a) and (b) respectively. The x-axis represents the focal length of the virtual camera in pixels.}
  \label{fig:Intrinsic}
  \vspace{-4mm}
\end{figure}

\noindent\textbf{Intrinsic generalization}
To verify the robustness of methods to camera intrinsic change, we change the focal length of the cameras with fixed resolution. As shown in the Fig.~\ref{fig:Intrinsic} (a) and (b), focal length changes affect VideoPose, PoseFormer, RIE to varying degrees under MPJPE and MRPE metrics respectively. For instance, with only 4\% variation on focal length, MRPE of the baseline approaches increase more than 50\%. In contrast, both Ray3D and Ray3D\_w/o\_CE achieve stable result. This result clearly showcases the merits of our ray-based input representation. 

\noindent\textbf{Extrinsic generalization}
To evaluate the impact of the change of extrinsics on generalizability, we change the rotation, pitch angles and translation of the camera pose respectively. Note that the translation is measured by the euclidean distance between camera and subject. In addition, we design a new baseline approach for root joint localization. Specifically, we estimate the height of the root joint using the mean average height from subjects of H36M, \ie, $93.95$cm. Using this height assumption we may localize the root joint along its 3D ray. We term this approach as Ray Fixed Root Height (RFRH). 

\begin{figure}
  \centering
  \includegraphics[width=0.44\linewidth]{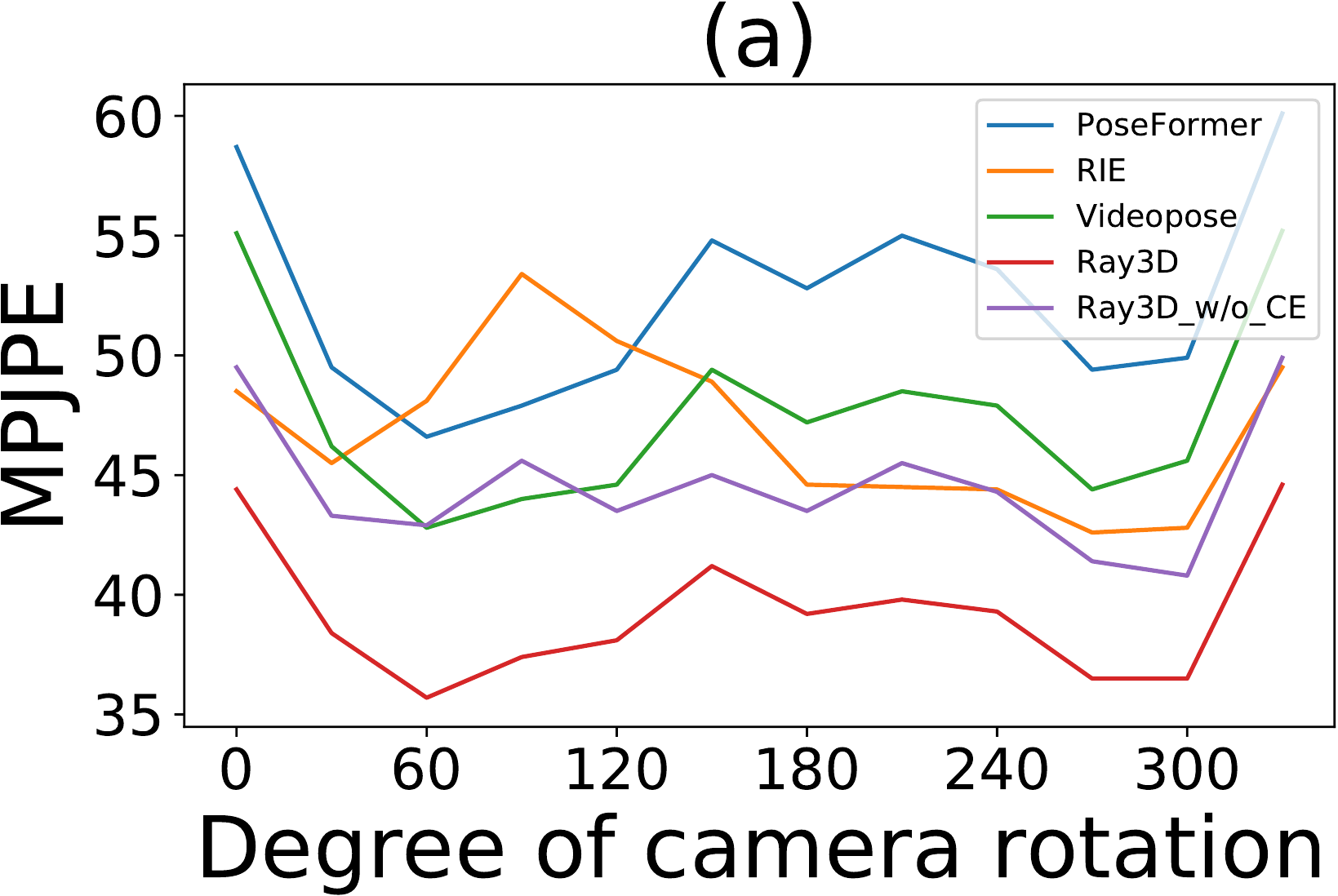}
  \includegraphics[width=0.44\linewidth]{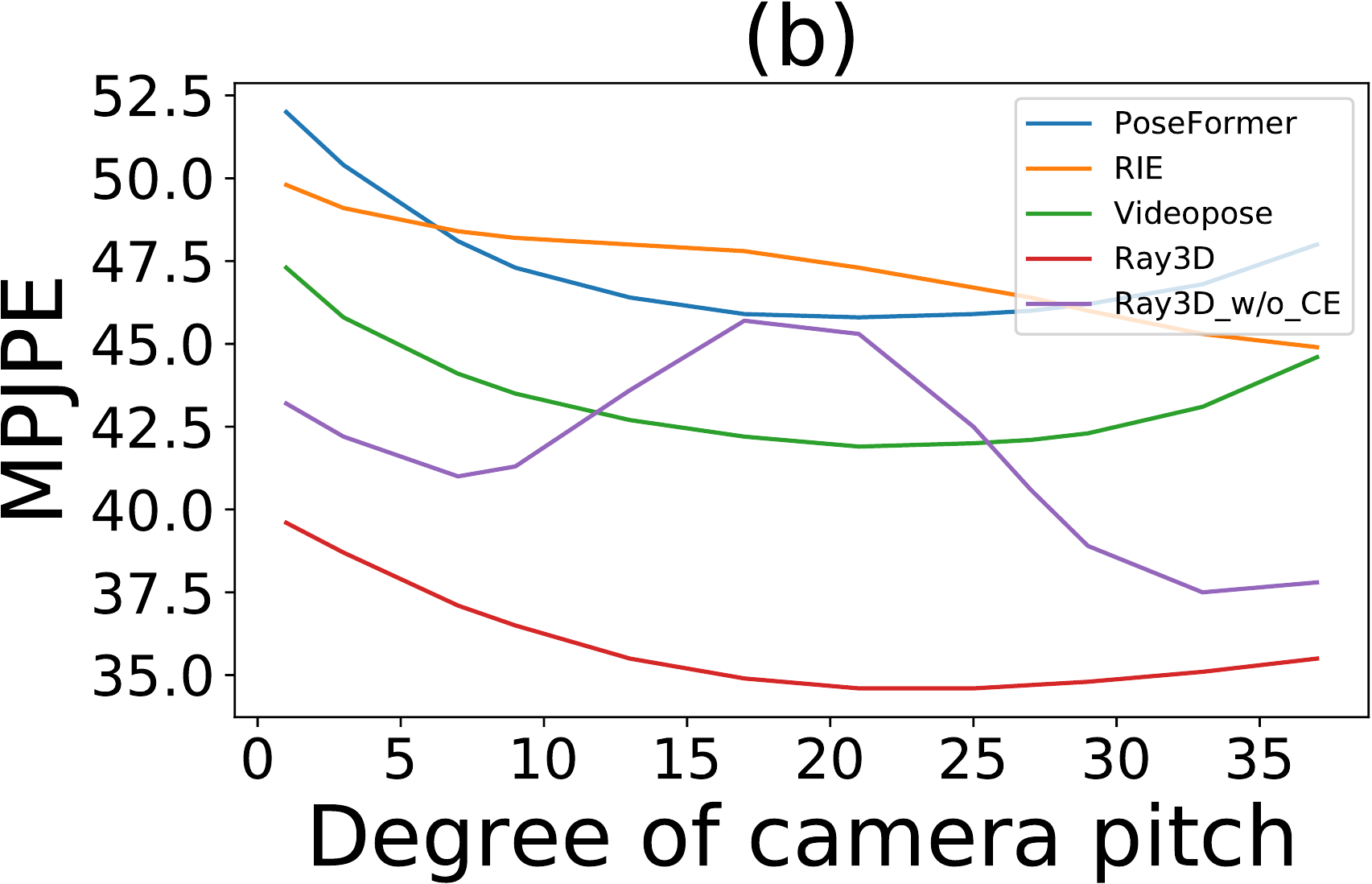}
  \includegraphics[width=0.44\linewidth]{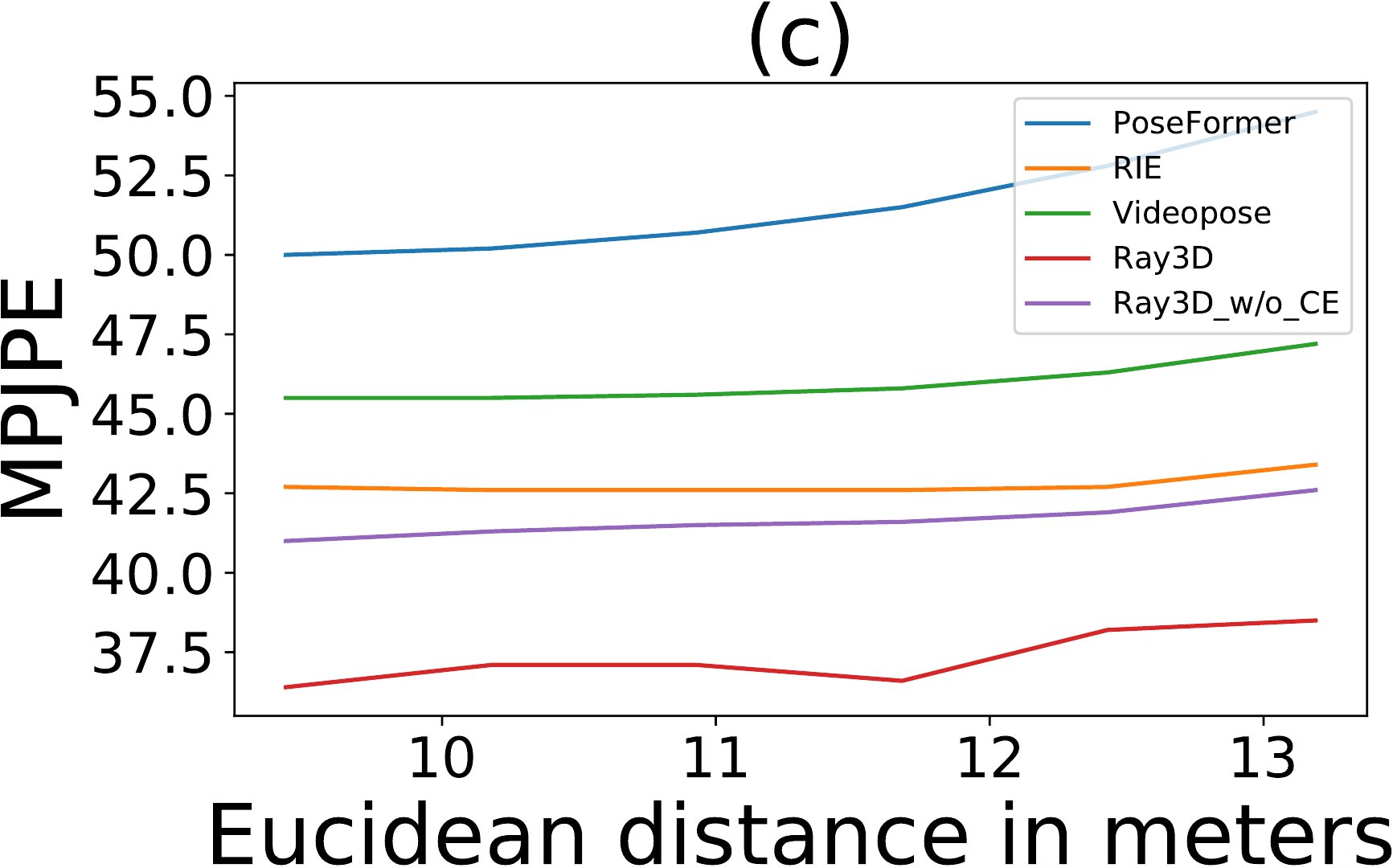}
  \includegraphics[width=0.44\linewidth]{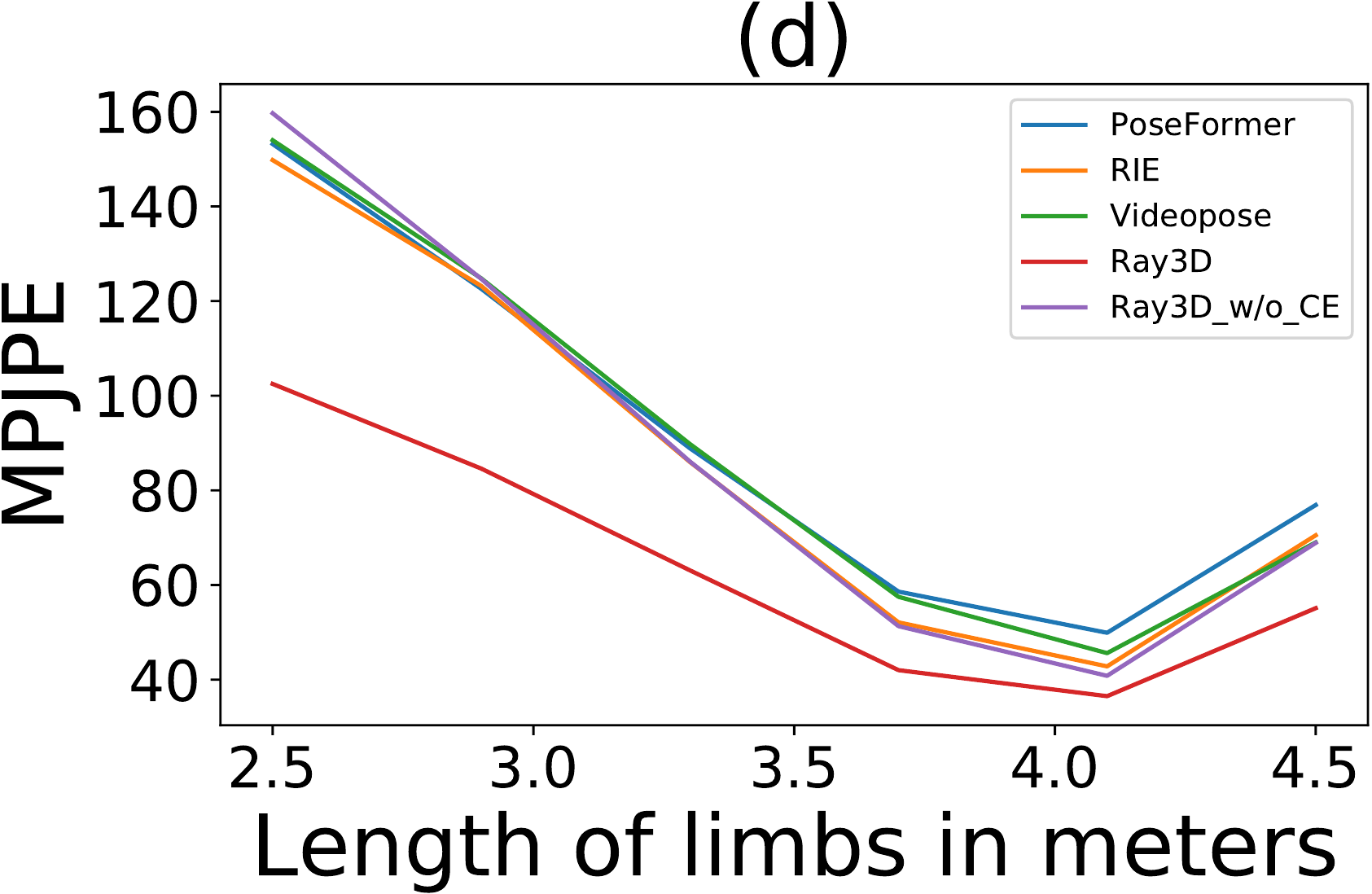}
  \vskip-9pt
  \caption{Figures (a), (b), (c) and (d) showcase the performance using MPJPE metric in case of rotation, camera pitch, translation and body scale variations correspondingly. The x-axis denotes the degree of camera rotation, the degree of camera pitch, euclidean distance between camera and subject in meters and the total length of human limbs in meters respectively.}
  \label{fig:Extrinsic_1}
  \vspace{-2mm}
\end{figure}

\begin{figure}
  \centering
  \includegraphics[width=0.44\linewidth]{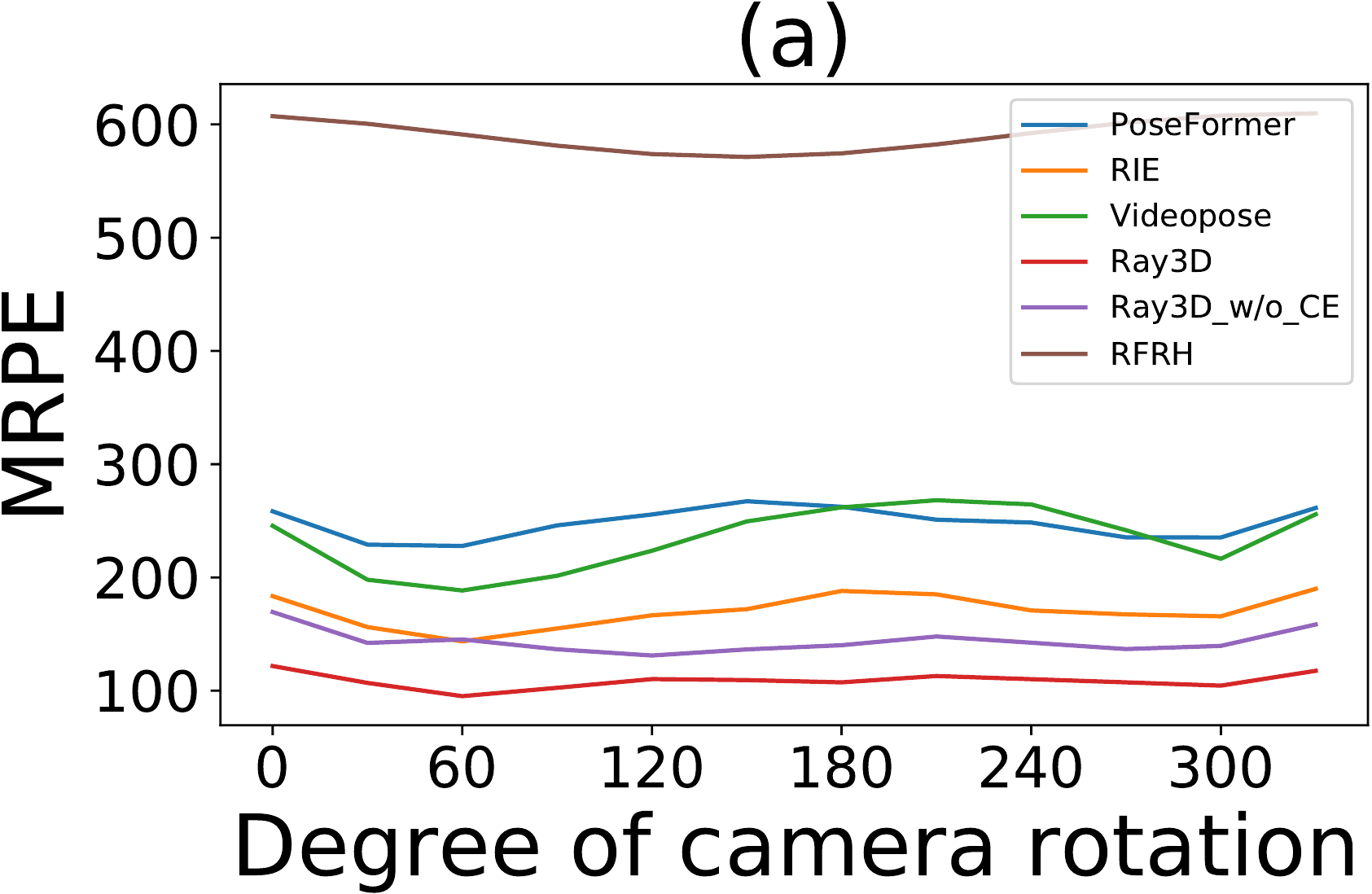}
  \includegraphics[width=0.44\linewidth]{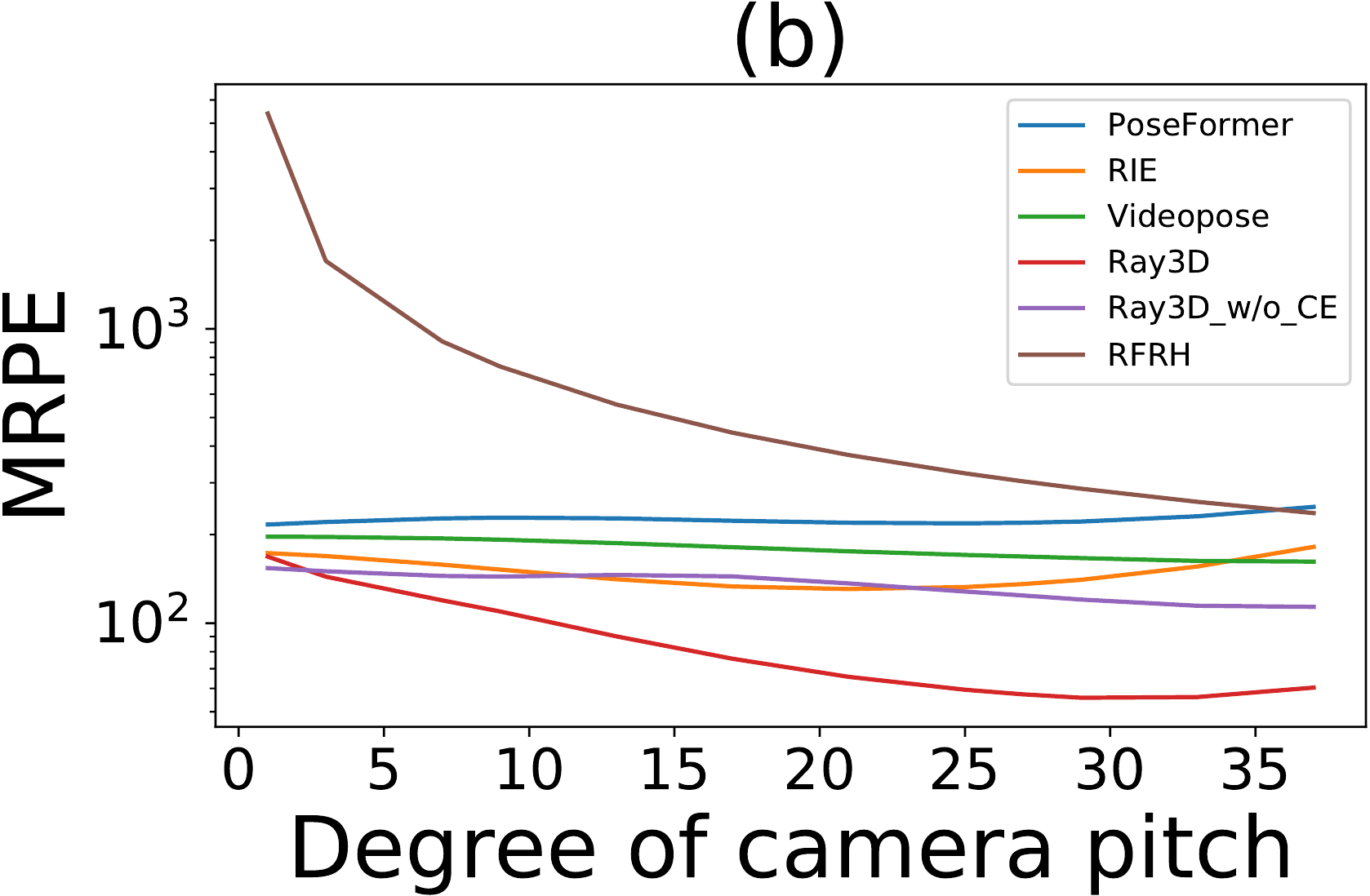}
  \includegraphics[width=0.44\linewidth]{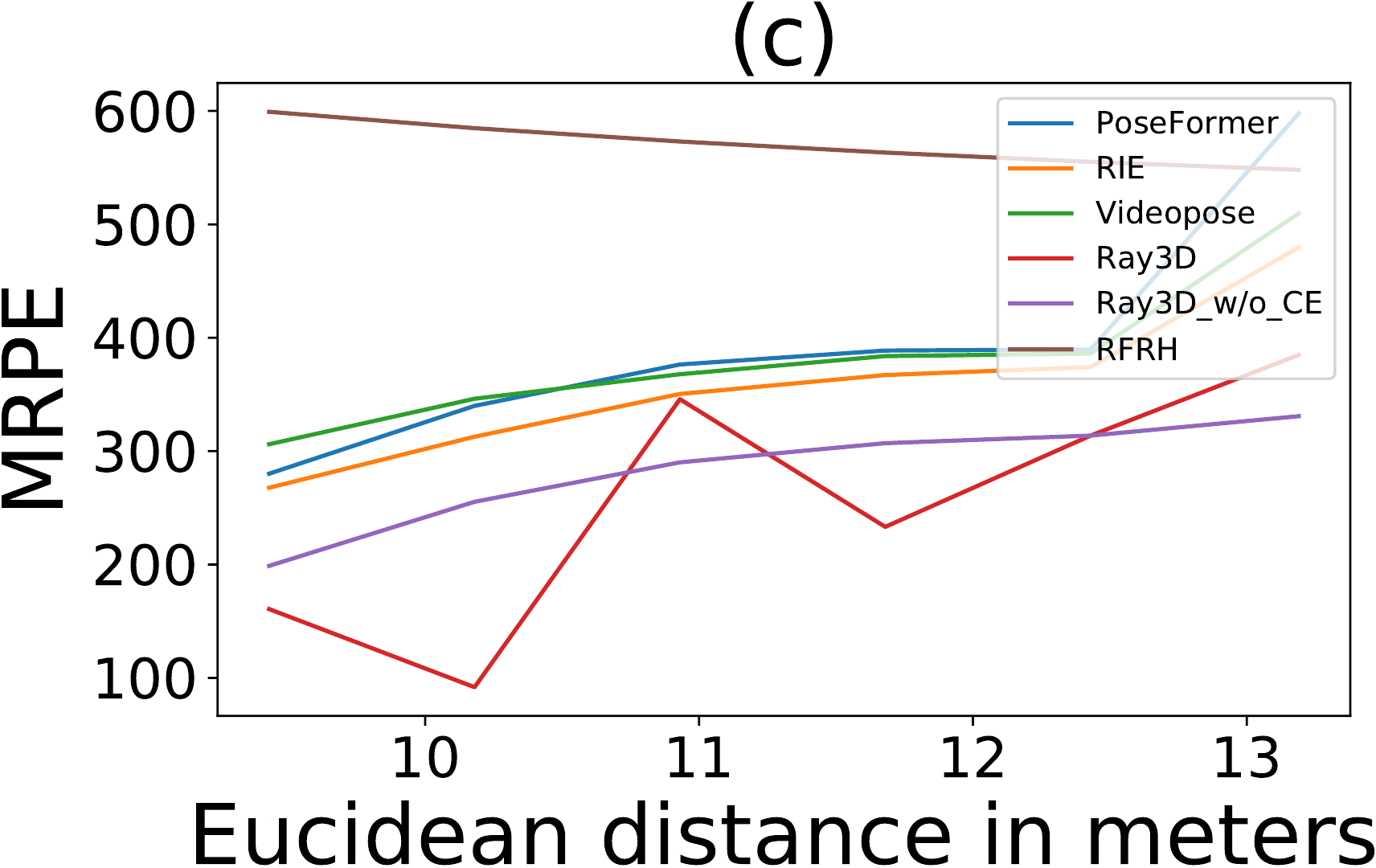}
  \includegraphics[width=0.44\linewidth]{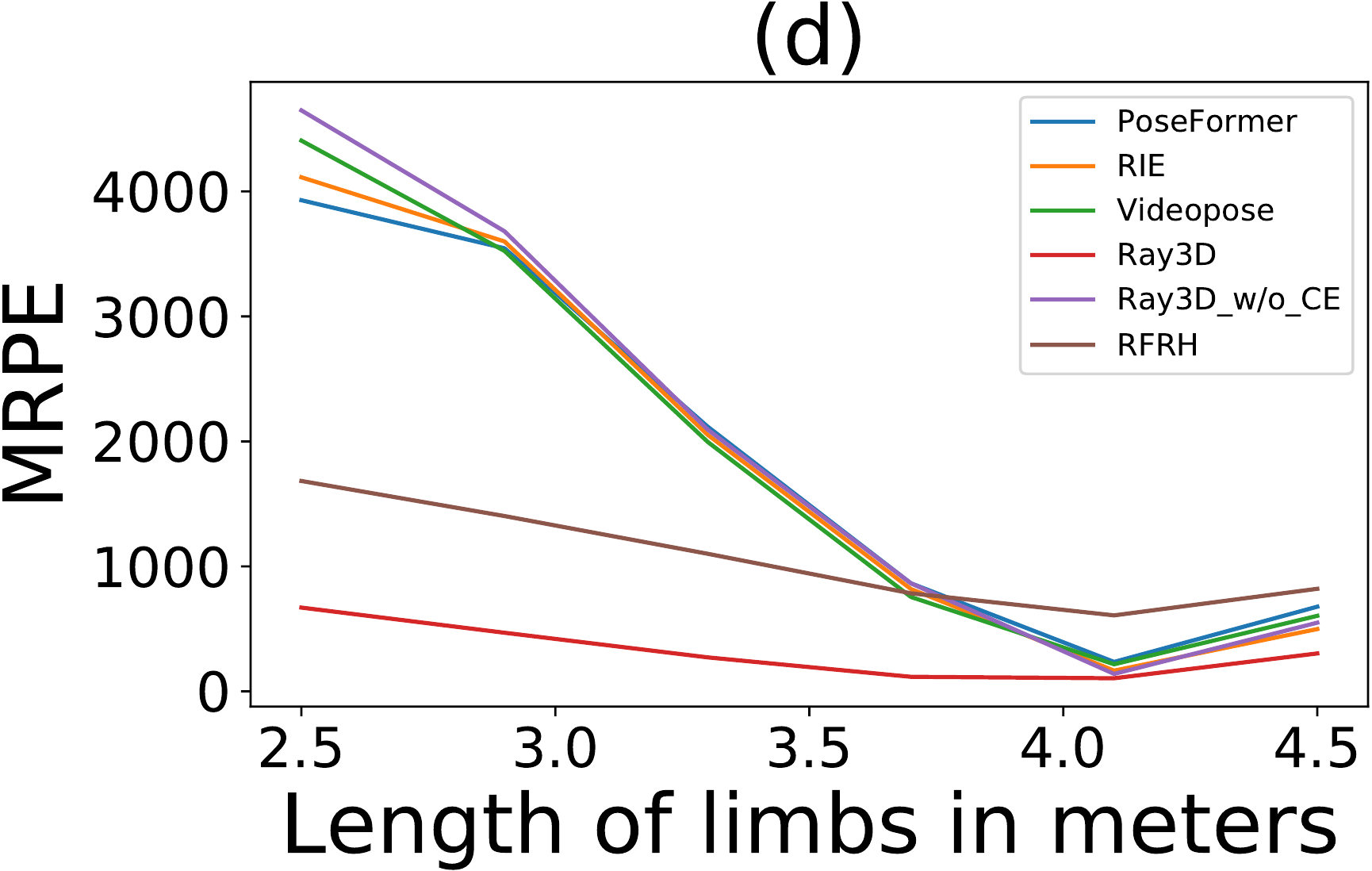}
  \vskip-9pt
  \caption{Figures (a), (b), (c) and (d) showcase the performance using MRPE metric in case of rotation, camera pitch, translation and body scale variations correspondingly. The x-axis denotes the degree of camera rotation, the degree of camera pitch, euclidean distance between camera and subject in meters and the total length of human limbs in meters respectively.}
  \label{fig:Extrinsic_2}
  \vspace{-4mm}
\end{figure}

To measure the robustness of model against rotation variations, we rotate the camera around the scene center while keeping the camera height and camera pitch the same. As shown in Fig.~\ref{fig:Extrinsic_1} (a) and Fig.~\ref{fig:Extrinsic_2} (a), Ray3D outperforms baseline methods by large margin on both MPJPE and MRPE, which indicates Ray3D is not only able to accurately localize the joint in WCS, it's also able to estimate the root-relative pose robustly. Ray3D\_w/o\_CE shows better results than RIE, which indicates that normalized ray representation works better than vanilla 2D keypoints. Among the baseline approaches, the learning-based approaches achieve better result than RFRH. This demonstrates the learning based methods indeed manage to resolve the ambiguities to some extent through data-driven way. RFRH performs poorly due to the violation of root height assumption in the evaluation dataset. For instance, when the subject sits down on the floor, the root joint height can be close to 0.

To evaluate robustness against camera pitch variation, we change pitch angle of the cameras to various degrees while keeping the same distance between camera and the subject. As shown in Fig.~\ref{fig:Extrinsic_1} (b) and Fig.~\ref{fig:Extrinsic_2} (b), Ray3D consistently outperforms baselines at all pitch angles for MRPE and MPJPE. 

Similarly, to evaluate model robustness against camera translation, we generate a batch of cameras with constant pitch angle and gradually changing distance from the subject. As shown in Fig.~\ref{fig:Extrinsic_1} (c) and Fig.~\ref{fig:Extrinsic_2} (c), baseline methods suffer from performance degradation as the camera moves away from subjects. In contrast, Ray3D achieves satisfactory results.

\noindent\textbf{Person scale generalization}
To verify robustness against the person scale ambiguity, we change the bone length of H36M body poses to 0.6-1.1 times as done in PoseAug~\cite{GongZF21}. The experimental results are shown in Fig.~\ref{fig:Extrinsic_1} (d) and~\ref{fig:Extrinsic_2} (d). Accuracy of all the comparing approaches degrades notably when the body size is small. For instance, for the smallest body figure, MRPE of PoseFormer, RIE, Videopose reach 4 meters, which is even higher than rule-based RFRH. MRPE of Ray3D increases to 800mm, which is still much better than baselines.

Throughout these systematic experiments on the synthetic dataset, we can safely arrive following conclusions. By converting 2D keypoints to normalized rays, both our Ray3D and Ray3D\_w/o\_CE achieve stable and accurate performance regardless of camera intrinsics variations. By adding camera embedding, Ray3D outperforms Ray3D\_w/o\_CE with clear margin for most of the testing cases except for a few camera settings in Fig.~\ref{fig:Extrinsic_2} (c). This verifies the effectiveness of camera embedding for both root-relative pose estimation and root joint absolute localization.  
\subsection{Ablation Studies}
\label{subsec: exp_4_5}
In this experiment, we further study the effectiveness of intrinsic decoupling, camera coordinate normalization and camera embedding using cross-dataset setting. We use RIE network as baseline model and add our modules gradually. All the models are trained on H36M dataset and tested on 3DHP dataset with 17-keypoint 2D poses as input. Table~\ref{table:ablation} summarizes the results. As we can see, MRPE keeps decreasing when adding proposed modules. Intrinsic decoupling improves the baseline model with a large margin. Camera normalization also drives the model to be more generalizable. With camera embedding, the trajectory estimation produces the best MRPE.

\begin{table}[htbp]
\centering
\small
\caption{Ablation study on 3DHP dataset with models trained with H36M dataset. IND denotes intrinsic decoupling, Nor. represents camera normalization. CE is camera embedding. We use RIE as base model. Best results are shown in \textbf{bold}. }
\vskip-12pt
\begin{tabular}{@{}l|lll|ll@{}}
\midrule  
   Method        & IND            & Nor.          & CE          & MRPE               \\ \midrule      
   RIE           &\XSolid         &  \XSolid      & \XSolid     & 1079.2             \\ 
   RIE\_w\_IND     &\Checkmark      &  \XSolid      & \XSolid     & 448.9              \\
   Ray3D\_w/o\_CE  &\Checkmark      &  \Checkmark   & \XSolid     & 311.6              \\  
   Ray3D         &\Checkmark      &  \Checkmark   & \Checkmark  & \textbf{307.4}     \\ \bottomrule     
\end{tabular}
\label{table:ablation}
\end{table}

\subsection{Discussion}
\label{exp: discuss}
Our Ray3D approach outperforms the baseline methods significantly in terms of generalizability both on the three realistic benchmarks and on the synthetic dataset. This clearly showcases the robustness of the Ray3D approach. However, Ray3D's performance drops when the body figure size varies a lot, as shown in Fig.~\ref{fig:Extrinsic_2} (d). This is mainly due to the fact that all the training body poses are adult.  Meanwhile, our approach assumes the subject is on the ground plane. The model may fail if the subject is off the ground for a long time (\eg., climbing a ladder). Additionally, calibrated camera parameters need to be provided, which limits the use cases of Ray3D. Accurate 3D pose estimation might be misused for surveillance applications where skeleton configuration estimation can assist person identification. We advocate proper usage.

\section{Conclusion}
In this paper, we present an innovative monocular absolute  3D human pose estimation approach named Ray3D. This approach gradually resolves the inherent ambiguities through a series of novel designs: conversion from 2D keypoints to 3D normalized rays; temporal fusion of 3D rays; inclusion of camera embedding. As a result, Ray3D significantly outperforms SOTA methods on three realistic benchmarks and one synthetic benchmark. 

\clearpage
{\small
\bibliographystyle{ieee_fullname}
\bibliography{egbib}
}

\clearpage
\appendix   

\section*{Supplementary material}

\section{Calculation of camera height and camera pitch}
In this section, we introduce ways of calculation of camera height and pitch. We assume that the WCS is constructed with respect to the constraint that axis $x$ and axis $y$ of the WCS are placed on the plane ground. 

\subsection*{Camera height}
The 3D coordinates of the perspective camera $M=[X_W, Y_W, Z_W]^{T}$ in the WCS is calculated with its own rotation matrix ${R_{W2C}}$, and translation vector ${T_{W2C}}$ as follows: 
\begin{equation}
    M = -R_{W2C}^{-1}\cdot{T_{W2C}}.
    \label{eq:cam_coord_world}
\end{equation}
Such that the camera height equals to $Z_W$. 

\subsection*{Camera pitch}
Define the unit vector along the camera optical axis as $\boldsymbol{V_C}=[0, 0, 1]^{T}$ in the CCS. One can easily calculate its representation $\boldsymbol{V_W}$ in the WCS as: 
\begin{equation}
    \boldsymbol{V_W} = R_{C2W}\cdot{\boldsymbol{V_C}}.
    \label{eq:cam_ray_world}
\end{equation}
Then  it is straightforward to compute camera pitch ${\theta}$ that defines the angle between the optical axis of the camera and the ground plane in the WCS.

\section{Synthetic dataset generation}
\label{sec:augmentation}
To systematically evaluate the robustness of 3D human pose estimators against variations of camera intrinsic and extrinsic parameters, we create a synthetic dataset with intrinsic and extrinsic parameters augmentation from H36M dataset. In this section, we describe how we generate this synthetic dataset in details. First, the mathematics formulation of intrinsic/extrinsic augmentation is introduced. Then we describe the adopted augmentation parameters.  

\subsection*{Intrinsic parameters augmentation} 
\label{subsec: intrinsci_aug}
For camera intrinsic parameters augmentation, focal length $f$ and principal points $c$ are modified while keeping the projected 2D keypoints within the field of view. The augmentation is performed as follows:
\begin{equation}
    \begin{aligned}
        &\hat{f} = f + \Delta f, \\
        &\hat{c} = c + \Delta c, \\
        &\begin{array}{r@{\quad}l}
             s.t. &0 \leq \{x_i\}_{i=1}^{J} \leq W_{\emph{I}},  \\
             &0 \leq \{y_i\}_{i=1}^{J} \leq H_{\emph{I}}
        \end{array}.
    \end{aligned}
\end{equation}

${H_{\emph{I}}}$ and ${W_{\emph{I}}}$ as the height and width of the image frame. ${x_{i}}$ and ${y_{i}}$ are the projected 2D keypoint location. Different camera intrinsics result in different 2D keypoints in the image. Without 3D ray representation, the network struggles to predict the same 3D pose facing camera intrinsic variation.

\subsection*{Extrinsic parameters augmentation} 
\label{subsec: extrinsci_aug}
For camera extrinsic parameters augmentation, we modify camera viewpoint $\beta$ (camera rotation), the relative distance between subject and the camera $\gamma$ (camera translation) and pitch of the camera $\theta$ (camera pitch). The augmentation is conducted as follows with constraint that the projected 2D keypoints are in the image frame:  
\begin{equation}
    \begin{aligned}
        &\hat{\beta} = \beta + \Delta \beta, \\
        &\hat{\gamma} = \gamma + \Delta \gamma, \\
        &\hat{\theta} = \theta + \Delta \theta, \\
        &\begin{array}{r@{\quad}l}
             s.t. &0 \leq \{x_i\}_{i=1}^{J} \leq W_{\emph{I}},  \\
             &0 \leq \{y_i\}_{i=1}^{J} \leq H_{\emph{I}}
        \end{array}.
    \end{aligned}
\end{equation}

By changing camera rotation, camera translation and camera pitch, we may generate various located virtual cameras. Camera embedding is learned for every camera for generalisation of lifting network, which is helpful for accurate trajectory prediction. The specific augmentation parameters are detailed in the following section.

\begin{table*}[ht]
\centering
\tiny
\caption{Technical summary of synthetic dataset constructed based on H36M.}
\begin{tabular}{l|llllll|l}
\toprule
                  Dataset & Num. of camera pose  & focal length/pixel  & x-coordinate of principal point/pixel &camera rotation/degree  &camera pitch/degree &camera translation/meter & subjects           \\ \midrule
training          & 324                  & [1143:1150]         &[508:514]                            &[60:300:120]            &[2:38:2]            & [9.05:11.70:0.76]       & S1, S5, S6, S7, S8 \\
extrinsic testing & 126                  & [1143:1150]         &[508:514]                            &[0:360:30]              &[1:37:2]            & [9.43:13.19:0.76]       & S9, S11            \\
intrinsic testing & 100                  & [1100:1180]         &[450:550]                            &0                       &12                  & 4.5                     & S9, S11            \\ \bottomrule
\end{tabular}
\label{table:dataset}
\\
\end{table*}

\subsection*{Augmentation parameters}
Without loss of generality, we randomly select $1$ camera whose id number is \emph{55011271} from H36M to conduct camera augmentation. The overall summary of the dataset is shown in Table~\ref{table:dataset}. 

To evaluate model performance against camera intrinsic variations, we generate the intrinsic testing dataset. Specifically, the focal length and the coordinate of principal points\footnote{We set the same value for x coordinate and y coordinate of principal point for simplicity.} of simulated cameras are augmented. Note that for training dataset and extrinsic testing set, camera intrinsics are the same as original set-up in H36M. For instance, the focal length of simulated cameras in the intrinsic testing set ranges from $1100$ to $1180$, compared with the cameras from training set whose focal length is in the $1143$-$1150$ range. Similarly, for the ${x}$ coordinate of principal point of simulated cameras, it has a longer range of $450$ to $550$ in the intrinsic testing dataset, compared with a range of $508$ to $514$ in the training dataset. In total, we have $100$ virtual cameras generated with fixed extrinsic for intrinsic generalization test. 

For extrinsic generalization test, camera rotation, camera pitch and camera translation are augmented. Specifically, camera rotation ranges from $0$ to $360$ degrees at $30$ degree interval such that extrinsic-testing cameras evenly revolve around the subjects. Camera pitch ranges from $0$ to $40$ degrees, which covers both frontal-view camera and large-pitch cameras. The interval of camera pitch is 2 degrees for both training dataset and extrinsic testing dataset. Camera translation ranges from $9$ to $14$ meters such that the relative distance between camera and subject is changing from the near to the distant. The interval of camera translation is $0.76$ meter for training and extrinsic testing cameras. In total, $126$ virtual cameras are generated with fixed intrinsic for extrinsic generalization test. And $324$ cameras are generated for training, such that camera embedding module learns to cope with vast range of camera pose variations. We set the augmentation parameter to the range which are common in the real-world scenarios (\eg., unmanned stores).

As for person scale generalization, the total length of augmented human limbs (bone length) ranges from $2.5$ to $4.5$ meters with the height of human ranging from $1$ meter to $2$ meters correspondingly. Note that the synthetic 3D human skeletons are only used for person scale generalization test, but excluded during model training stage.

\section{Supplementary  experiments}
In this section, we report additional  evaluation results to fully analyze the proposed Ray3D on public and synthetic datasets.

\begin{table*}[htbp]
\centering
\tiny
\caption{Quantitative evaluation results under MPJPE on H36M using detected keypoints as input. (f = 9) means this approach utilizes 9 consecutive frames for pose estimation, and (f = 1) means the approach does not make use of temporal information. * means this approach using 2D keypoints detected by CPN. Best results are shown in \textbf{bold}.}
\begin{tabular}{@{}l|llllllllllllllll|l@{}}
\toprule
Detected keypoints as input     &                 & Dir.   & Disc. & Eat.  & Greet & Phone & Photo & Pose  & Purch. & Sit   & SitD. & Somke & Wait  & WalkD.& Walk  & WalkT. & Average.        \\ \midrule
Dabral et al. \cite{dabral2018learning}           &ECCV'18 & 44.8  & 50.4  & 44.7  & 49.0  & 52.9  & 61.4  & 43.5   & 45.5  & 63.1  & 87.3  & 51.7  & 48.5  & 52.2  & 37.6   & 41.9   & 52.1   \\
Cai et al. (f = 7) \cite{cai2019exploiting}       &ICCV'19 & 44.6  & 47.4  & 45.6  & 48.8  & 50.8  & 59.0  & 47.2   & 43.9  & 57.9  & 61.9  & 49.7  & 46.6  & 51.3  & 37.1   & 39.4   & 48.8   \\
Videopose. (f = 9)* \cite{dario2019videopose}     &CVPR'19 & 46.4  & 48.9  & 45.7  & 49.8  & 52.5  & 61.5  & 47.7   & 46.8  & 59.9  & 68.1  & 50.7  & 47.5  & 52.7  & 38.4   & 42.1   & 50.6   \\
Yeh et al.\cite{yeh2019chirality}                 &NIPS'19 & 44.8  & 46.1  & \textbf{43.3}   & 46.4  & 49.0  & \textbf{55.2}   & 44.6   & 44.0  & 58.3  & 62.7  & 47.1  & 43.9  & 48.6  & 32.7   & 33.3   & 46.7   \\
UGCN  (f = 96) \cite{wang2020motion}              &ECCV’20 & \textbf{41.3}  & \textbf{43.9}   & 44.0  & \textbf{42.2}   & \textbf{48.0}   & 57.1  & \textbf{42.2}    & \textbf{43.2}   & 57.3  &\textbf{61.3}   & \textbf{47.0}   & \textbf{43.5}   & \textbf{47.0}   & \textbf{32.6}    & \textbf{31.8}   & \textbf{45.6}   \\
PoseFormer (f = 9)* \cite{ce2021poseformer}       &ICCV'21 & 47.9  & 51.5  & 49.3  & 50.8  & 53.7  & 58.6  & 49.5   & 46.6  & 62.0  & 70.3  & 52.6  & 49.3  & 53.8  & 40.5   & 43.0   & 52.0   \\
PoseAug (f = 1)* \cite{GongZF21}                  &CVPR'21 & -     & -     & -     & -     & -     & -     & -      & -     & -     & -     & -     & -     & -     & -      & -      & 52.9   \\
RIE (f=9)* \cite{wenkang2021improving}            &ACMMM'21& 44.8  & 47.9  & 46.1  & 47.4  & 50.4  & 57.6  & 45.7   & 44.6  & \textbf{57.0}   & 64.2  & 49.5  & 45.7  & 50.9  & 36.6   & 39.8   & 48.6   \\ \hline
Ray3D (f = 9)*                                    &        & 44.7  & 48.7  & 48.7  & 48.4  & 51.0  & 59.9  & 46.8   & 46.9  & 58.7  & 61.7  & 50.2  & 46.4  & 51.5  & 38.6   & 41.8   & 49.7   \\ \bottomrule
\end{tabular}
\label{table:supp_result_1}
\\
\end{table*}

\begin{table*}[htbp]
\centering
\tiny
\caption{Quantitative evaluation results under Abs-MPJPE and MRPE on H36M using GT as 2D input. (f = 9) means this approach utilizes 9 consecutive frames for pose estimation, and (f = 1) means the approach does not make use of temporal information. Best results are shown in \textbf{bold}.}
\begin{tabular}{@{}l|llllllllllllllll|l@{}}
\toprule
Abs-MPJPE              &                                   &Dir.  &Disc. &Eat.  &Greet  &Phone & Photo & Pose & Purch.   &Sit     &SitD.  & Somke & Wait  & WalkD. & Walk & WalkT. & Average \\ \midrule
Videopose (f = 9)~\cite{dario2019videopose} &CVPR'19 &73.7  &99.6  &88.8  &82.8   &\textbf{81.7}  & 121.8 & 89.8 & 83.8     & 110.6  & 234.4 & 95.8  & 92.4  & 91.2   & 69.7 & 64.2   & 98.7    \\ 
PoseLifter (f = 1)~\cite{ju2019absposelifter}           &ICCV'19 &65.5  &86.9  &103.9 &81.4   &95.2  & 109.2 & 80.1 & 107.3    & 152.4  & 245.0 & 106.2 & 95.6  & 115.5  & 87.1 & 69.8   & 106.8   \\
PoseFormer (f = 9)~\cite{ce2021poseformer}        &ICCV'21 &88.3  &88.3  &91.3  &94.3   &96.1  & 127.8 & 101.0& 120.0    & 114.5  & 227.7 & 102.4 & 110.8 & 97.2   & 99.1 & 91.1   & 111.6   \\
RIE (f = 9)~\cite{wenkang2021improving}          &ACMMM'21&75.4  &90.7  &\textbf{80.5}  &80.9   &75.3  & 100.4 & 85.9 & 92.2     & \textbf{93.1}   & 200.9 & \textbf{86.5}  & 87.9  & 88.5   & 67.8 & 58.6   & 91.0    \\ \hline
Ray3D (f = 1)                                     &        &\textbf{60.2}  &75.2  &102.0 &70.6   &92.5  & \textbf{85.2}  & 71.7 & \textbf{67.5}     & 123.9  & 129.5 & 87.0  & 77.6  & 92.7   & 74.0 & 67.7   & 85.2    \\ 
Ray3D (f = 9)                                     &        &65.6  &\textbf{70.4}  &100.1 &\textbf{64.1}   &92.0  & 86.6  & \textbf{65.6} & 73.2     & 119.2  & \textbf{117.4} & 92.9  & \textbf{70.1}  & \textbf{77.1}   & \textbf{64.4} & \textbf{61.4}   & \textbf{81.4}    \\ \bottomrule
MRPE                    &                                  &Dir.  &Disc. &Eat.  &Greet  &Phone &Photo  & Pose & Purch.   &Sit     &SitD.  & Somke & Wait  & WalkD. & Walk & WalkT. & Average \\ \midrule
Videopose (f = 9)~\cite{dario2019videopose} &CVPR'19  &57.6  &88.0  &77.2  &69.4   &74.2  &110.3  & 71.4 & 73.3     & 97.0   & 225.9 & 86.7  & 77.5  & 80.9   & 61.2 & 52.2   & 86.9    \\ 
PoseLifter (f = 1)~\cite{ju2019absposelifter}           &ICCV'19  &51.3  &75.6  &87.8  &67.9   &83.0  & 96.3  & 63.8 & 100.0    & 138.6  & 231.6 & 93.5  & 83.8  & 108.4  & 73.1 & 51.1   & 93.7    \\
PoseFormer (f = 9)~\cite{ce2021poseformer}       &ICCV'21  &63.2  &\textbf{63.2}  &77.4  &77.4   &84.3  &114.6  & 76.8 & 103.1    & 96.5   & 215.8 & 88.0  & 90.2  & 85.5   & 89.3 & 78.0   & 95.9    \\
RIE (f = 9)~\cite{wenkang2021improving}         &ACMMM'21 &60.6  &78.3  &\textbf{69.5}  &69.5   &\textbf{65.1}  & 90.6  & 68.3 & 81.5     & \textbf{79.1}   & 192.1 & \textbf{76.2}  & 73.6  & 80.2   & 59.5 & \textbf{48.1}   & 79.5    \\ \hline
Ray3D (f = 1)                                    &         &\textbf{45.4}  &63.4  &97.7  &57.5   &88.0  & \textbf{74.4}  & \textbf{53.4} & 59.4     & 116.9  & 119.1 & 79.8  & 60.9  & 85.5   & 64.8 & 56.1   & \textbf{74.9}    \\ 
Ray3D (f = 9)                                    &         &59.3  &65.4  &99.8  &\textbf{55.1}   &93.1  & 80.5  & 55.2 & \textbf{70.9}     & 116.4  & \textbf{104.6} & 89.9  & \textbf{59.8}  & \textbf{70.3}   & \textbf{56.8} & 52.4   & 75.3    \\ \bottomrule
\end{tabular}
\label{table:supp_result_2}
\end{table*}

\subsection{Evaluation on public benchmarks}
\noindent\textbf{H36M evaluation}
Table~\ref{table:supp_result_1} shows the performance of the methods that focus on root-relative pose estimation where detected 2D keypoints are taken as input. When the number of video frames taken as input are similar, we can observe that our Ray3D obtains comparable results compared to SOTA methods under MPJPE metric in Camera Coordinate System (CCS). MPJPE of Ray3D surpasses PoseFormer~\cite{ce2021poseformer} and Videopose~\cite{dario2019videopose} by 2.3mm and 0.9mm respectively, but Ray3D performs worse than RIE~\cite{wenkang2021improving} by 1.1mm. We argue that Ray3D is designed for absolute 3D pose estimation in World Coordinate System (WCS), such performance of root-relative pose estimation in CCS is acceptable.

Table~\ref{table:supp_result_2} shows the results for absolute pose estimation in WCS using GT 2D poses on H36M dataset. It can be seen that Ray3D outperforms all SOTA methods for both Abs-MPJPE and MRPE with clear margin. Compared with RIE, our method reduces Abs-MPJPE by 9.6mm and MRPE by 4.2mm respectively. These results demonstrate that Ray3D is effective and generates more accurate absolute 3D locations. 

\noindent\textbf{3DHP evaluation}
Table~\ref{table:supp_result_3} shows the results for absolute pose estimation in WCS using GT 2D poses on 3DHP dataset. One can observe that Ray3D outperforms all SOTA methods for both Abs-MPJPE and MRPE with clear margin. For instance, compared with PoseFormer~\cite{ce2021poseformer}, our method reduces Abs-MPJPE by 44.4mm and MRPE by 51.7mm respectively. 

\begin{table*}[htbp]
\centering
\caption{Quantitative evaluation results under MPJPE, Abs-MPJPE and MRPE on 3DHP using GT as 2D input. (f = 9) means this approach utilizes 9 consecutive frames for pose estimation, and (f = 1) means the approach does not make use of temporal information. Best results are shown in \textbf{bold}.}
\begin{tabular}{@{}l|lcc}
\toprule
method \textbackslash metric                                    & MPJPE        & Abs-MPJPE        & MRPE  \\ \midrule
Videopose (f = 9)~\cite{dario2019videopose}                     & 52.5         & 148.4            & 145.8 \\
PoseFormer (f = 9)~\cite{ce2021poseformer}                      & \textbf{40.8}& 147.8            & 147.5 \\
PoseLifter (f = 1)~\cite{ju2019absposelifter}                   & 78.2         & 143.6            & 129.1 \\
RIE (f = 9)~\cite{wenkang2021improving}                         & 47.4         & 140.8            & 141.0 \\
Ray3D (f = 1)                                                   & 48.4         & 118.2            & 114.0 \\
Ray3D (f = 9)                                                   & 46.0         & \textbf{103.4}   & \textbf{95.8} \\\bottomrule
\end{tabular}
\label{table:supp_result_3}
\end{table*}

\noindent\textbf{Cross-dataset testing}
We train comparing models on H36M dataset, and evaluate them using H36M, Humaneva-I and 3DHP. 14-joint definition is applied for all datasets during cross-dataset testing. For H36M and 3DHP, we remove mid spine, neck and chin keypoints. As for Humaneva-I, the thorax key-point is removed out of original 15 joints. As shown in Table~\ref{table:supp_result_4}, none of the baselines work well in cross-scenario situations while the Ray3D shows good generalization performance in H36M, Humaneva-I and 3DHP dataset. For instance, PoseFormer~\cite{ce2021poseformer} is able to predict better root-relative pose than Ray3D, but it struggles to predict precise root joint. And PoseLifter~\cite{ju2019absposelifter} fails to generalize to cross datasets, achieving inferior MRPE performance.

\begin{table*}[htbp]
\caption{Cross dataset evaluation. We adopt a 14-joint skeleton training on H36M, testing on H36M, Humaneva-I and 3DHP datasets.
MPJPE, Abs-MPJPE and MRPE are adopted. (f = 9) means this approach utilizes 9 consecutive frames for pose estimation, and (f = 1) means the approach does not make use of temporal information. The unit of all numbers is mm. The best results are in \textbf{bold}.}
\centering
\small
\begin{tabular}{@{}l|lcc|lll|lcc@{}}
\toprule
method \textbackslash datasets & \multicolumn{3}{c|}{H36M} & \multicolumn{3}{c|}{HumanEva-I} & \multicolumn{3}{c}{3DHP} \\
                                                        & MPJPE  & Abs-MPJPE  & MRPE   & MPJPE  & Abs-MPJPE  & MRPE    & MPJPE          & Abs-MPJPE  & MRPE   \\  \midrule
Videopose (f = 9)~\cite{dario2019videopose}             & 46.1   & 133.4      & 120.9  & 85.1   & 284.6      & 283.1   & 104.6          & 1262.8     & 1266.1 \\
PoseFormer (f = 9)~\cite{ce2021poseformer}              & 50.0   & 146.6      & 129.8  & \textbf{79.4}   & 260.7      & 253.9   & \textbf{101.9}          & 1313.6     & 1320.1 \\
PoseLifter (f = 1)~\cite{ju2019absposelifter}           & 56.9   & 147.4      & 135.1  & 3690.2 & 15170.6    & 16082.6 & 1180.9         & 6839.0     & 6899.4 \\
RIE (f = 9)~\cite{wenkang2021improving}                 & 41.5   & 136.4      & 125.1  & 82.0   & 272.9      & 285.1   & 102.4          & 1185.0     & 1187.0 \\
CDG (f = 1) ~\cite{WangSF20} & 52.0 & - & - & - & - & - & 111.9 & - & - \\
Ray3D (f = 9)                                          & \textbf{39.3}   & \textbf{106.8}      & \textbf{98.5}   & 81.5   & \textbf{121.5}      & \textbf{99.3}    & 108.1          & \textbf{422.4}      & \textbf{406.0}  \\ \bottomrule
\end{tabular}
\label{table:supp_result_4}
\end{table*}

\noindent\textbf{Evaluation with noisy cameras} To test the robustness of Ray3D when taking noisy cameras parameters as input, we add gaussian noise to intrinsic parameters (i.e., focal length and center points) and extrinsic parameters (i.e., rotation and translation) of H36M's cameras respectively.

\begin{figure}
  \centering
  \includegraphics[width=1\linewidth]{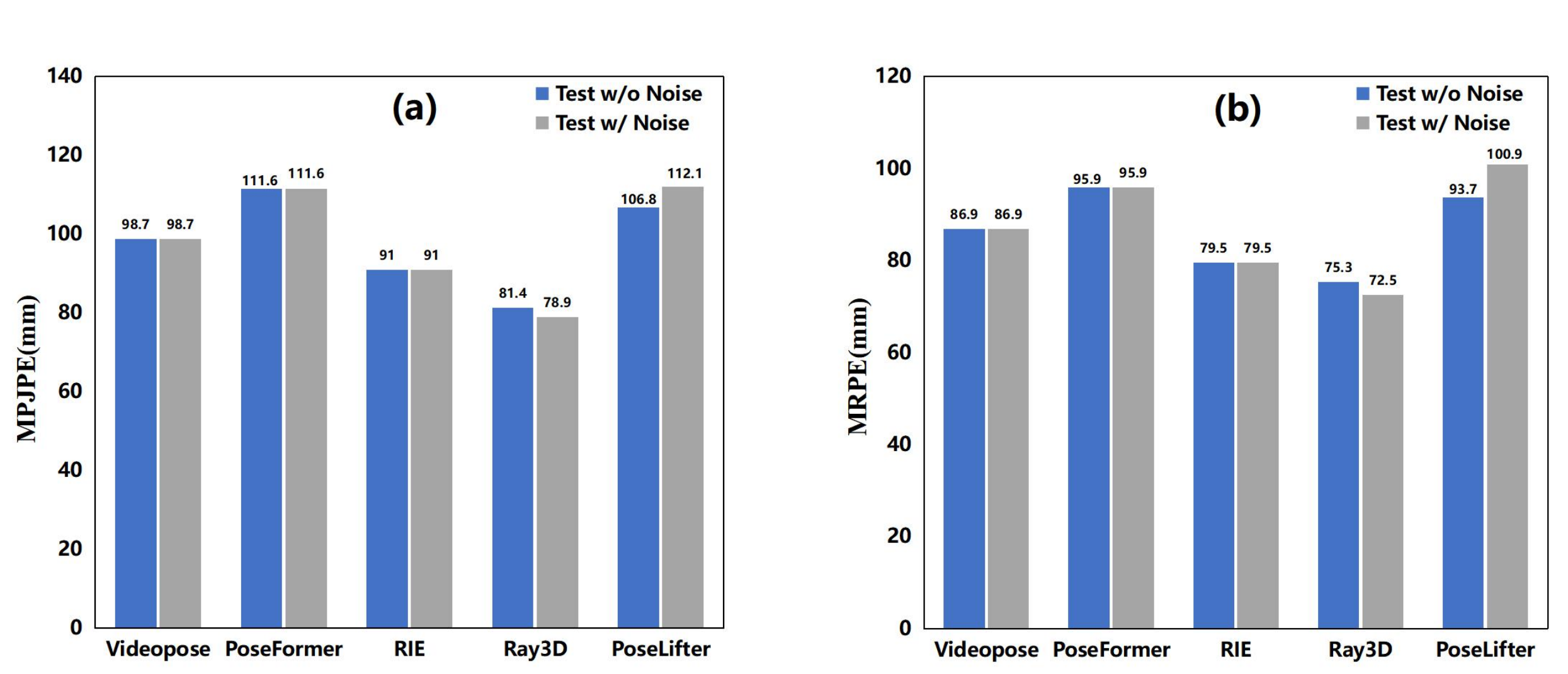}
  \caption{Performance under MPJPE and MRPE with noisy focal length are plotted in (a) and (b) respectively.}
  \label{fig:noisy_focal_length}
\end{figure}

\begin{figure}
  \centering
  \includegraphics[width=1\linewidth]{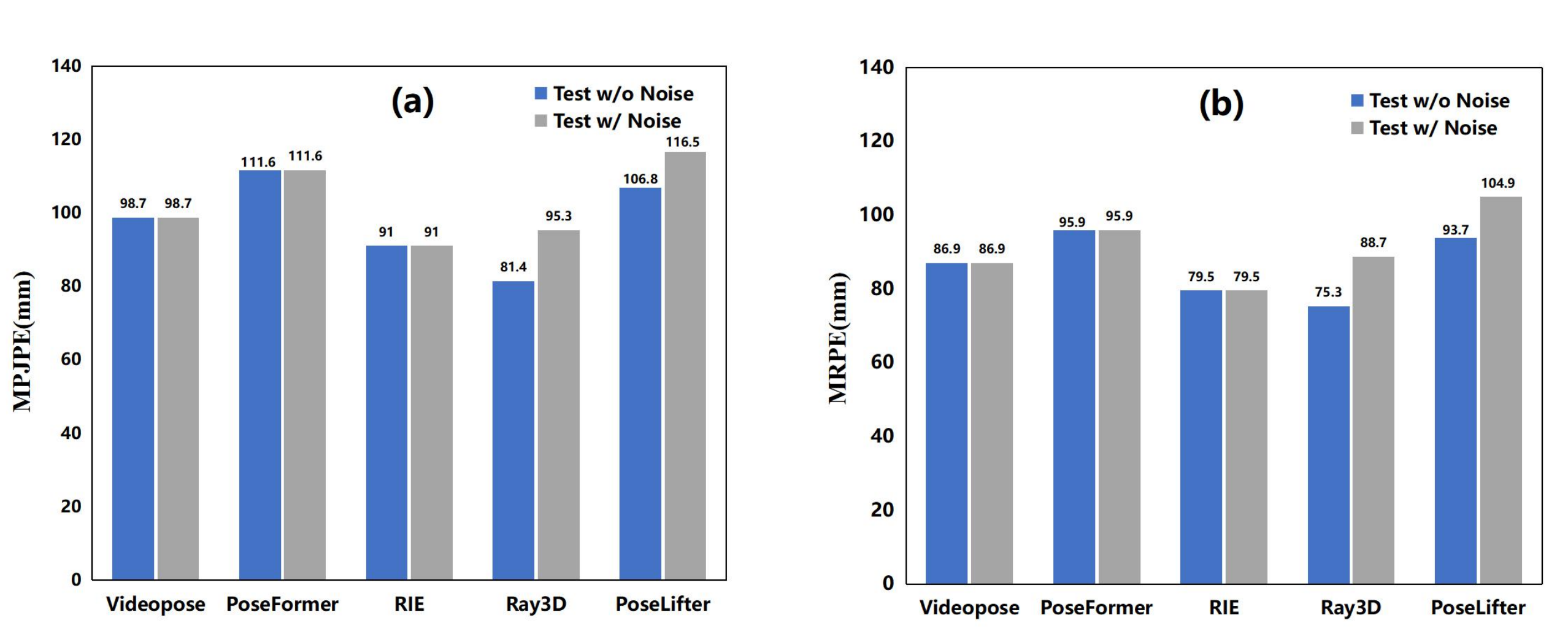}
  \caption{Performance under MPJPE and MRPE with noisy center points are plotted in (a) and (b) respectively.}
  \label{fig:noisy_center}
\end{figure}

The results of using noisy focal length and center points as input are shown in Fig.~\ref{fig:noisy_focal_length} and Fig.~\ref{fig:noisy_center}. As for the intrinsic parameters, Videopose~\cite{dario2019videopose}, PoseFormer~\cite{ce2021poseformer} and RIE~\cite{wenkang2021improving} do not use focal length and center points as input, noisy intrinsic parameters has no impact on these methods. While Ray3D and Poselifter~\cite{ju2019absposelifter} explicitly decouple the intrinsic parameters from the input. Noisy intrinsic parameters cause inaccurate decoupling, which results in slight performance changes. 

\begin{figure}
  \centering
  \includegraphics[width=1\linewidth]{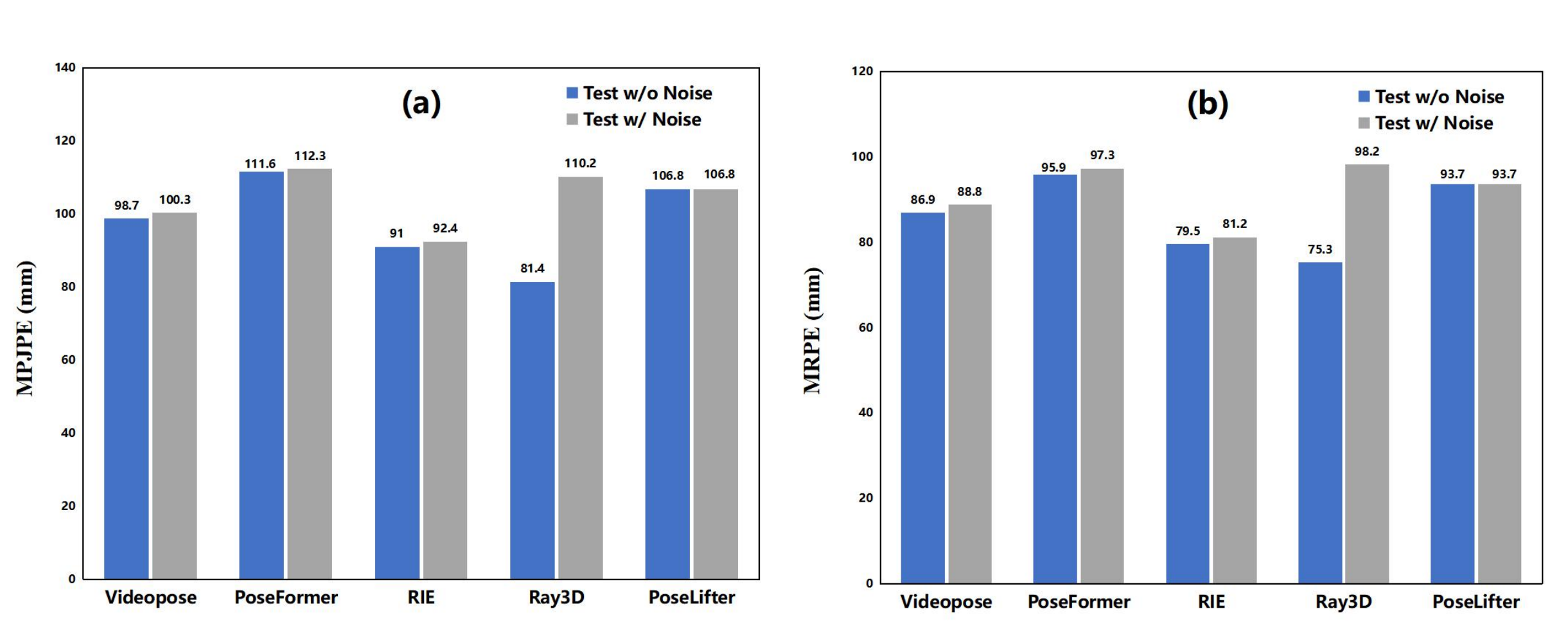}
  \caption{Performance under MPJPE and MRPE with noisy camera pitch are plotted in (a) and (b) respectively.}
  \label{fig:noisy_Rx}
\end{figure}

\begin{figure}
  \centering
  \includegraphics[width=1\linewidth]{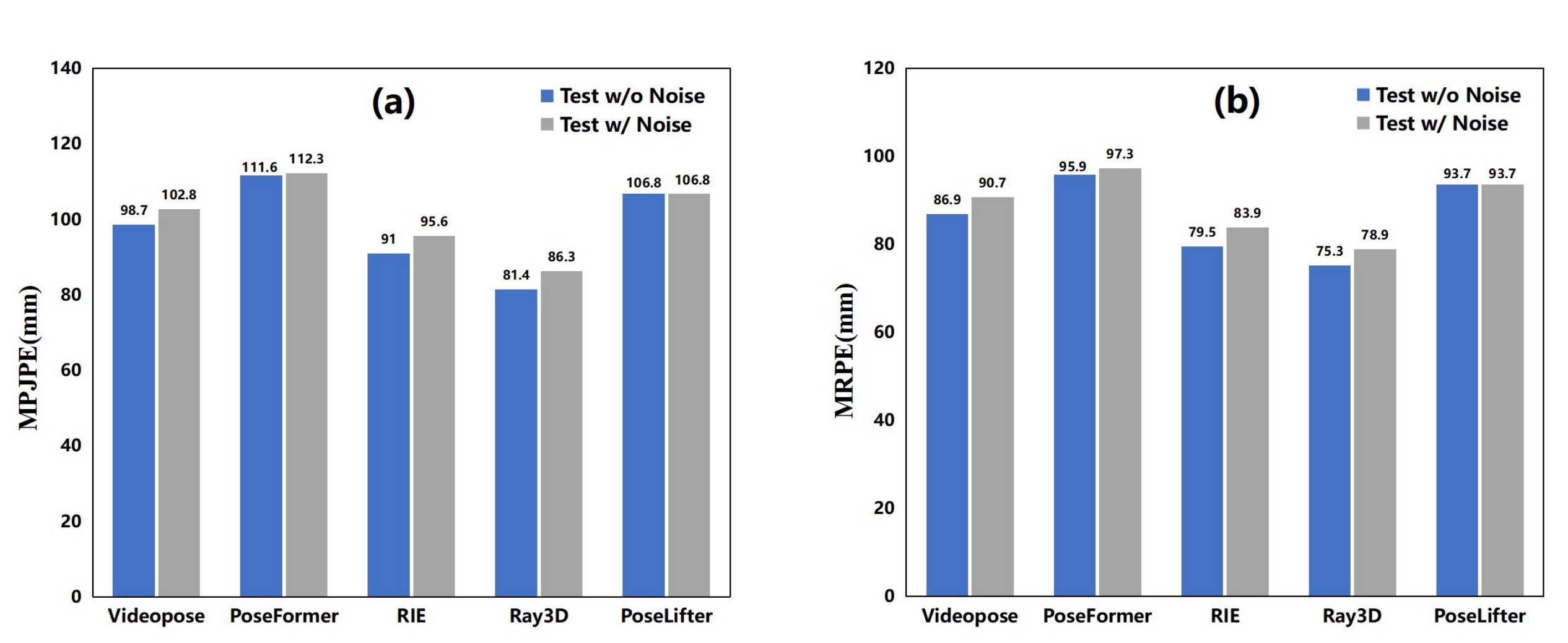}
  \caption{Performance under MPJPE and MRPE with noisy camera yaw are plotted in (a) and (b) respectively.}
  \label{fig:noisy_Ry}
\end{figure}

\begin{figure}
  \centering
  \includegraphics[width=1\linewidth]{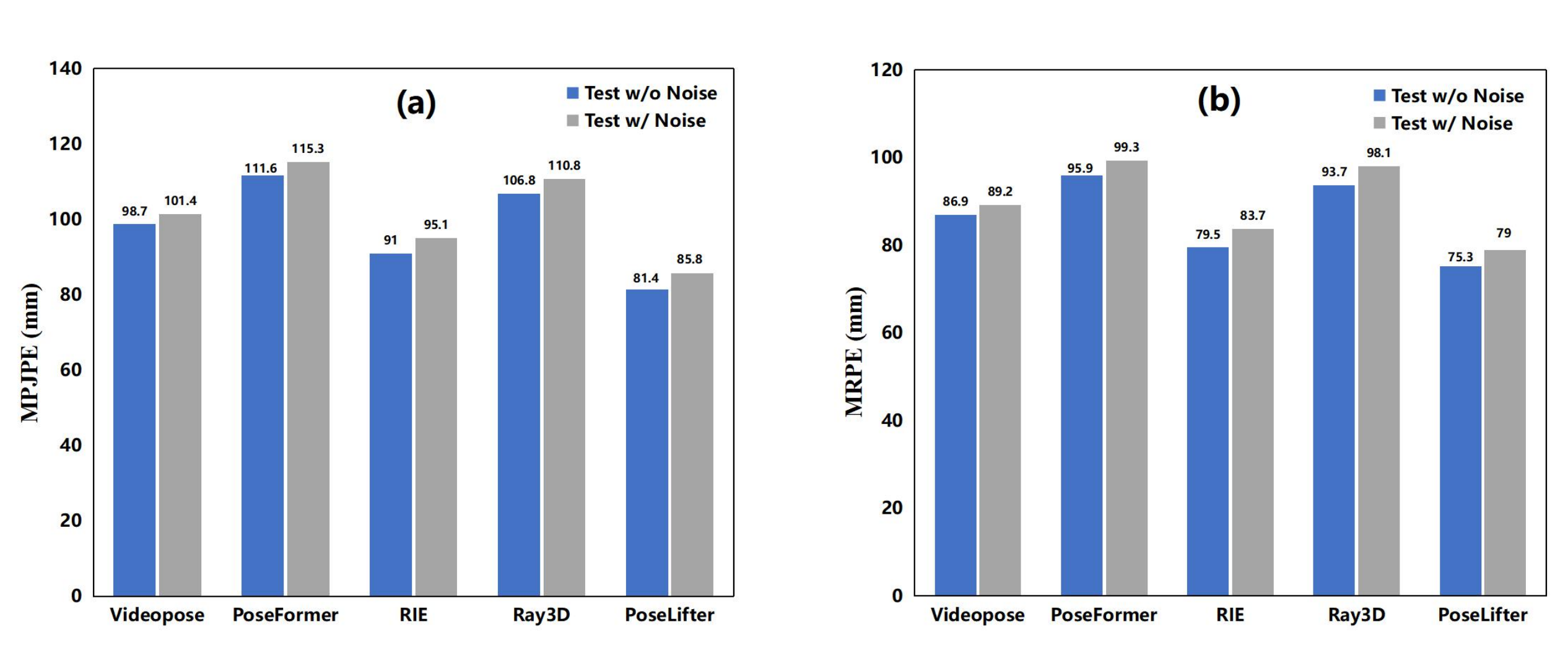}
  \caption{Performance under MPJPE and MRPE with noisy translation are plotted in (a) and (b) respectively. }
  \label{fig:noisy_Trans}
\end{figure}

As for the extrinsic parameters, Videopose~\cite{dario2019videopose}, PoseFormer~\cite{ce2021poseformer} and RIE~\cite{wenkang2021improving} use extrinsic parameters to convert final estimation from CCS to WCS, noisy extrinsic parameters cause performance degradation. Ray3D uses the well calibrated camera extrinsic parameters as an input, especially the camera height and camera pitch, which makes Ray3D sensitive to camera pitch change. As shown in the Fig.~\ref{fig:noisy_Rx}, after we added gaussian noise to camera pitch, the MPJPE increases from 81.2mm to 110.6mm. Fig.~\ref{fig:noisy_Ry} shows the performance with noisy camera yaw, Ray3D does not decrease significantly. As shown in Fig.~\ref{fig:noisy_Trans}, all methods have the same performance drop when provided with noisy camera translation.


\subsection{Evaluation on synthetic benchmarks}
\begin{figure}
  \centering
  \includegraphics[width=0.44\linewidth]{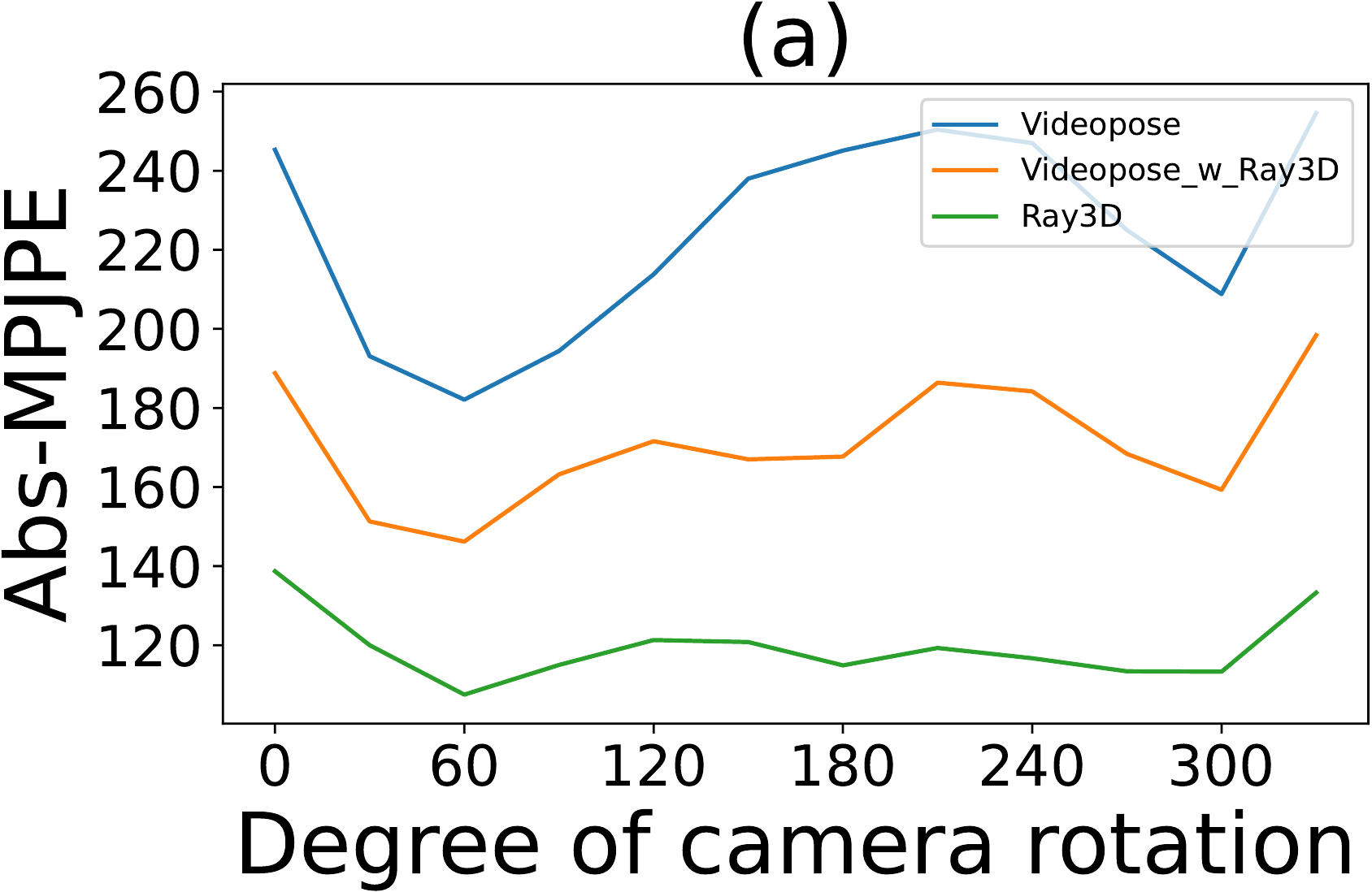}
  \includegraphics[width=0.44\linewidth]{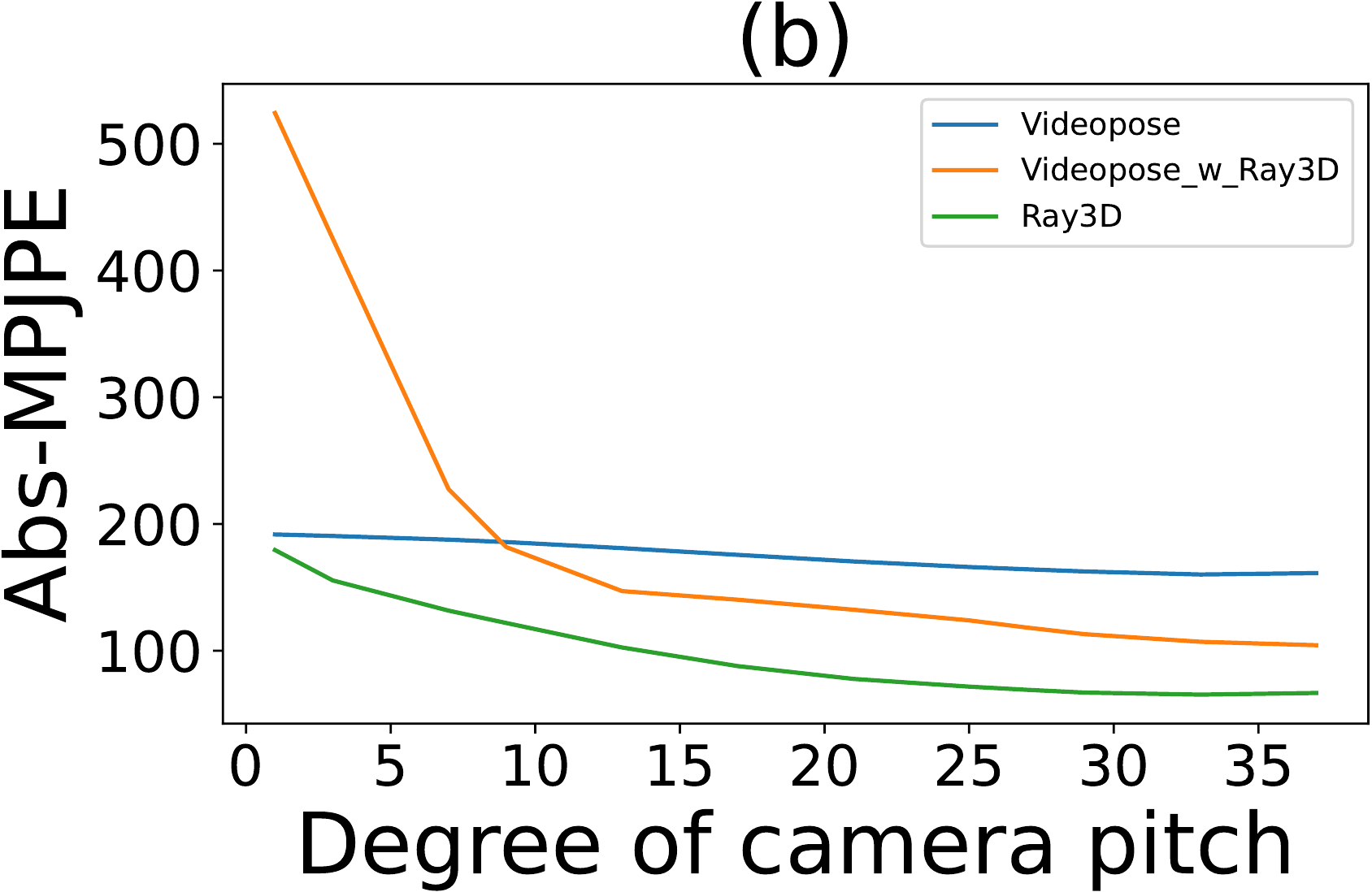}
  \includegraphics[width=0.44\linewidth]{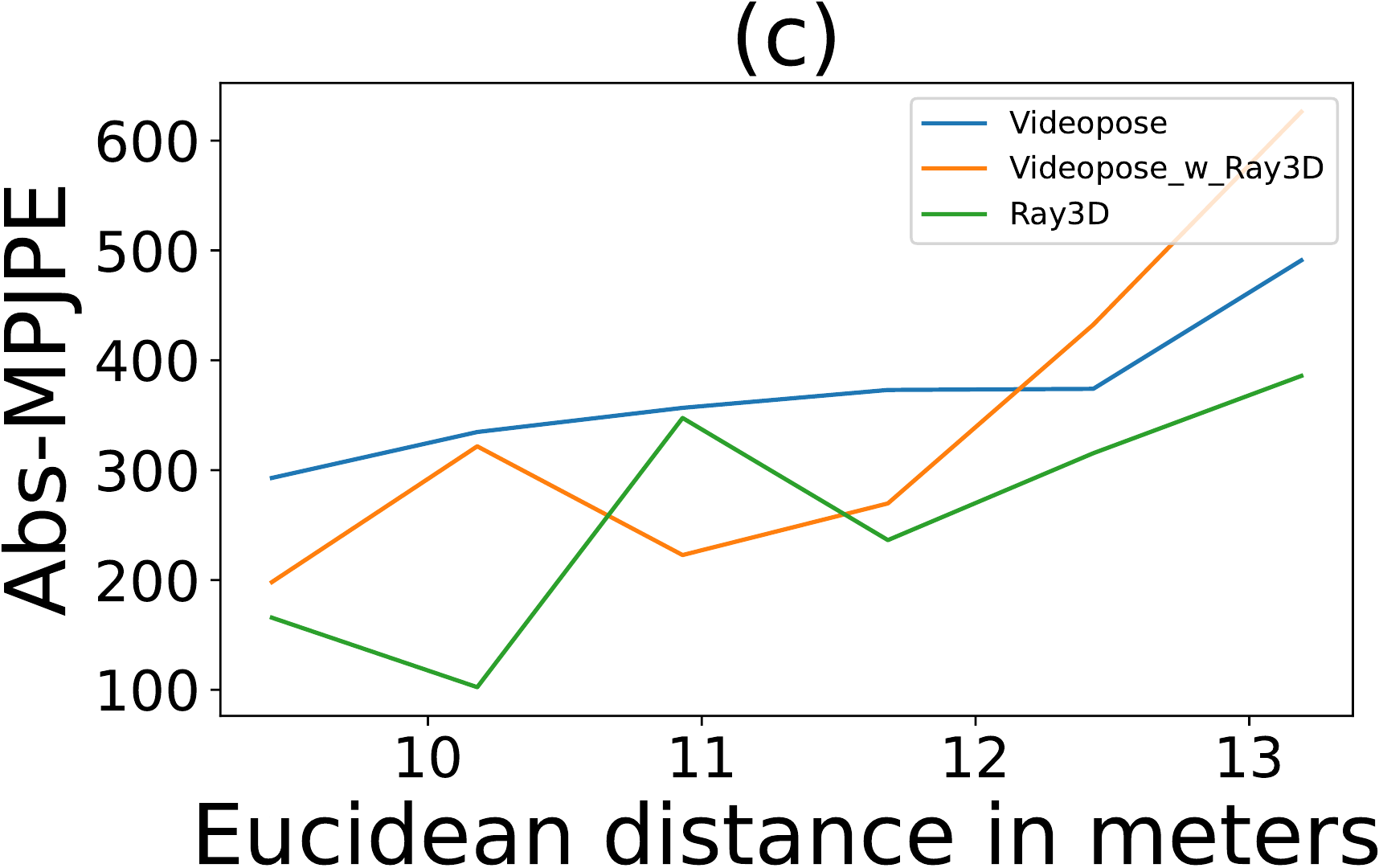}
  \includegraphics[width=0.44\linewidth]{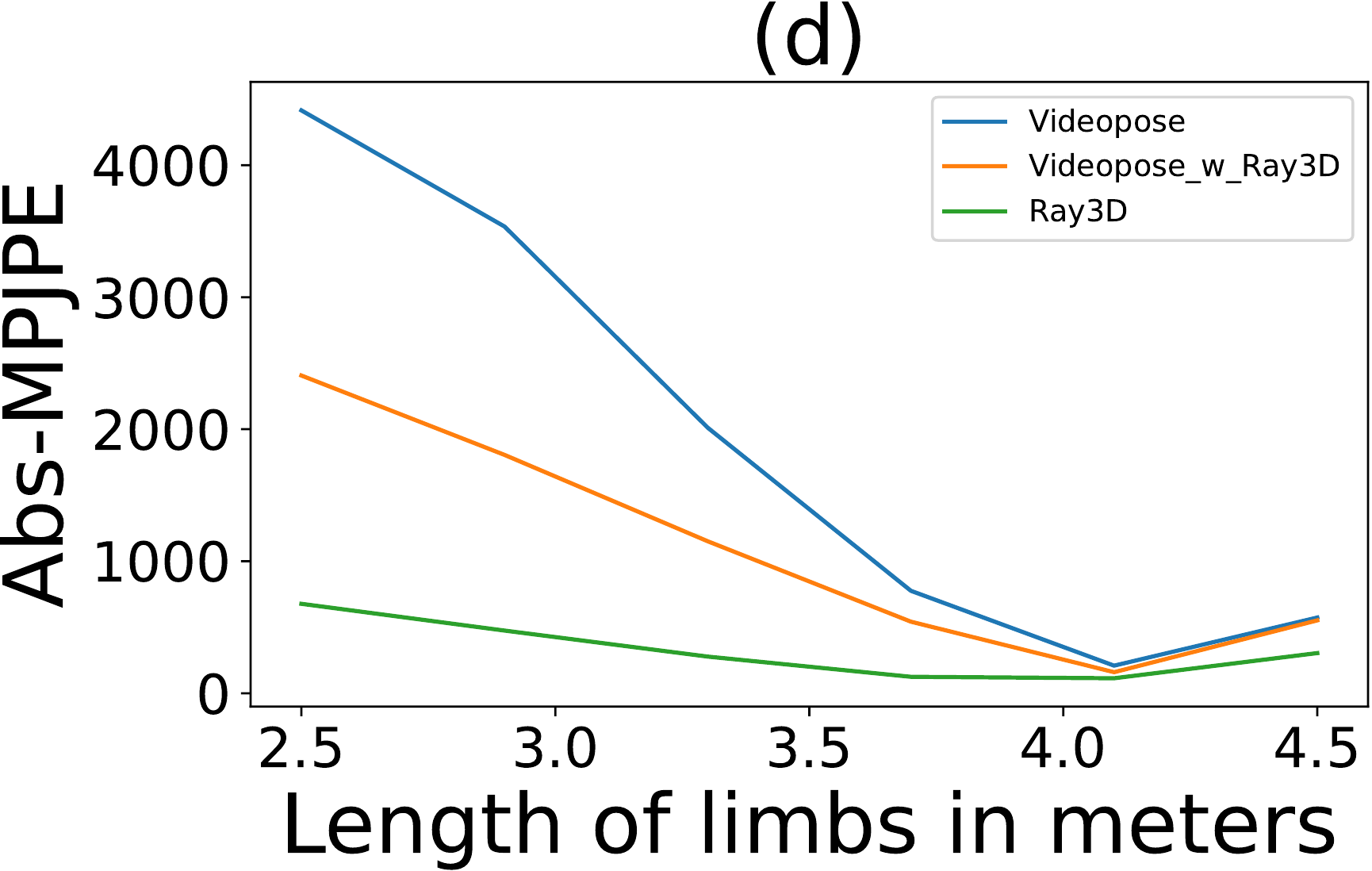}
  \caption{Figures (a), (b), (c) and (d) showcase the performance using Abs-MPJPE metric in case of rotation, camera pitch, translation and body scale variations correspondingly. The x-axis denotes the degree of camera rotation, the degree of camera pitch, euclidean distance between camera and subject in meters and the total length of human limbs in meters respectively.}
  \label{fig:Extrinsic_1_VP_Ray_3D}
\end{figure}

\begin{figure}
  \centering
  \includegraphics[width=0.44\linewidth]{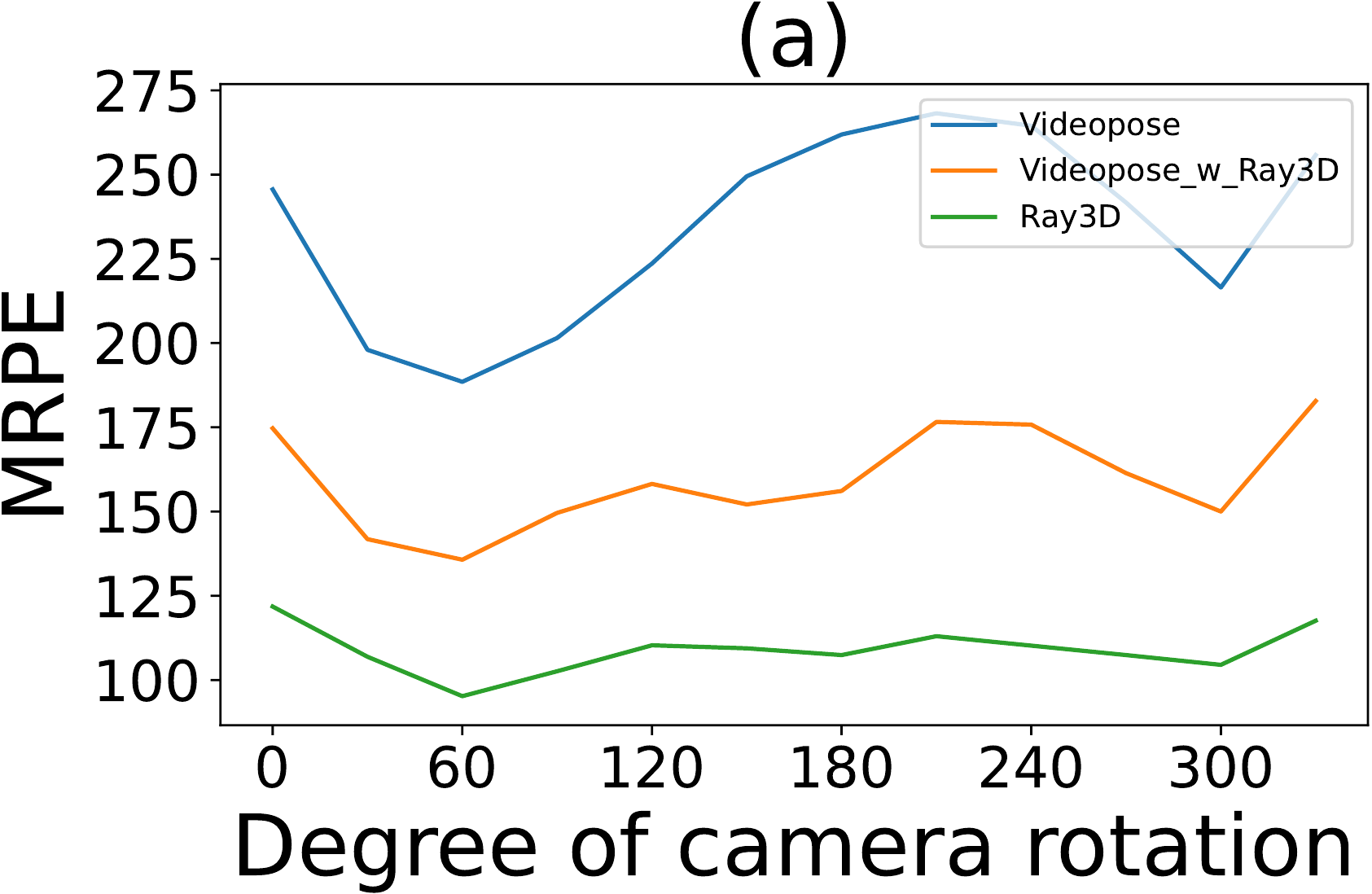}
  \includegraphics[width=0.44\linewidth]{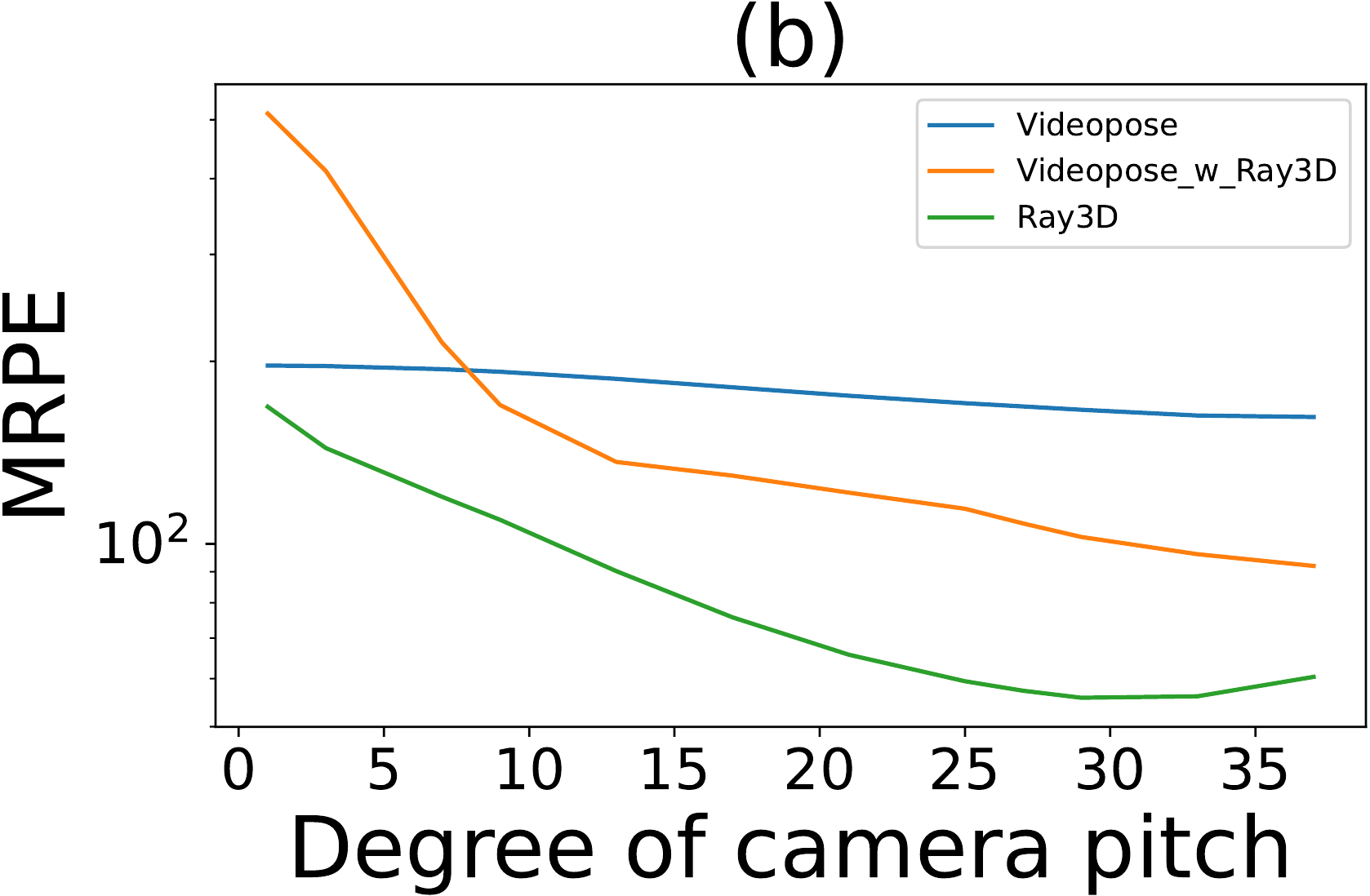}
  \includegraphics[width=0.44\linewidth]{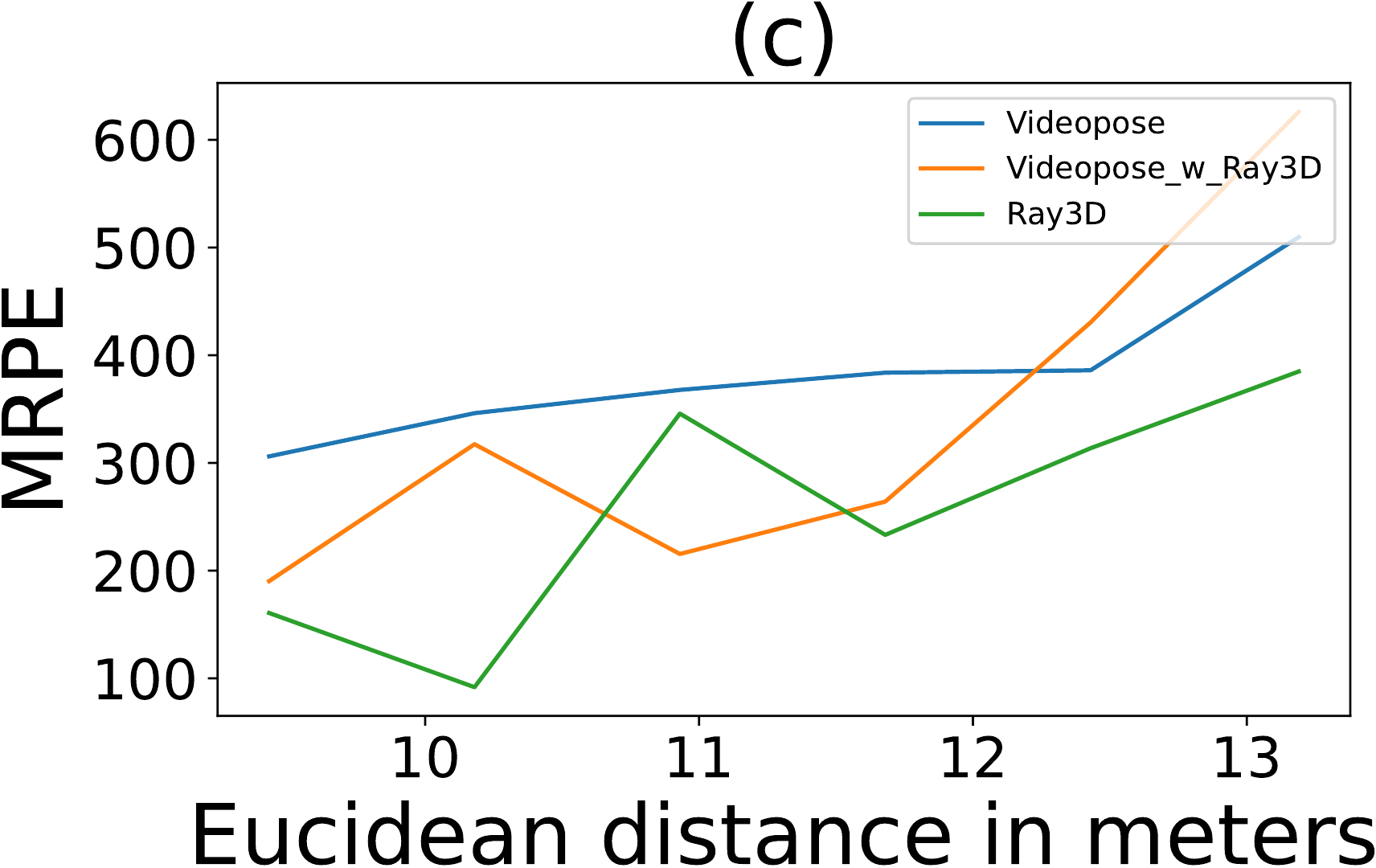}
  \includegraphics[width=0.44\linewidth]{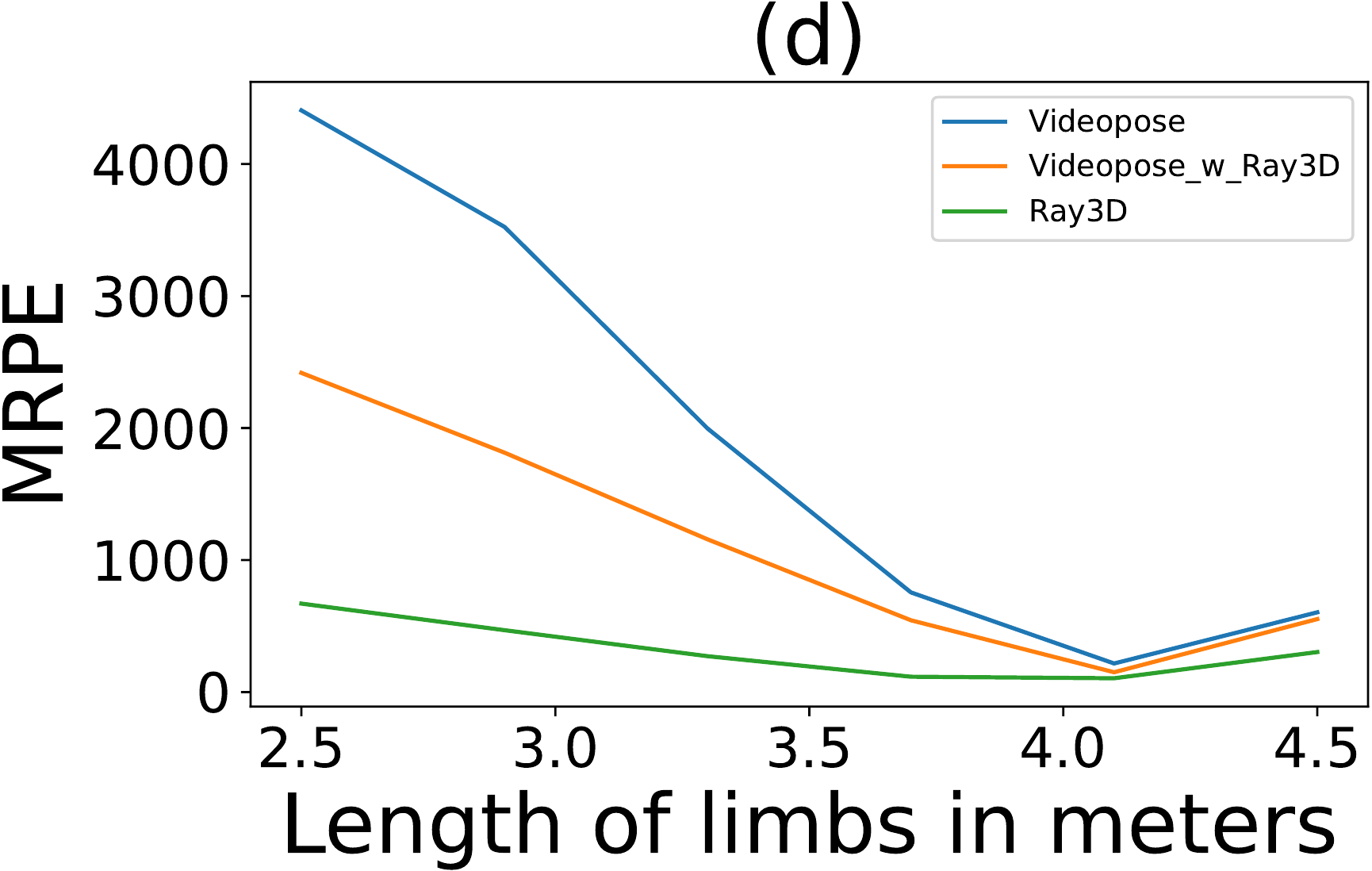}
  \caption{Figures (a), (b), (c) and (d) showcase the performance using MRPE metric in case of rotation, camera pitch, translation and body scale variations correspondingly. The x-axis denotes the degree of camera rotation, the degree of camera pitch, euclidean distance between camera and subject in meters and the total length of human limbs in meters respectively.}
  \label{fig:Extrinsic_2_VP_Ray_3D}
\end{figure}
\noindent\textbf{Integrate Videopose with Ray3D}
We integrate proposed Ray3D techniques with another baseline method Videopose~\cite{dario2019videopose}. The model is trained and evaluated on the proposed synthetic dataset. As shown in the Fig.~\ref{fig:Extrinsic_1_VP_Ray_3D}, Videopose~\cite{dario2019videopose} integrated with 3D ray representation and camera embedding techniques performs better than vanilla method under Abs-MPJPE metric. Same performance gain can be observed in the Fig.~\ref{fig:Extrinsic_2_VP_Ray_3D} under MRPE metric, which showcases that Ray3D incorporated to the different existing frameworks bring consistent improvement.

\begin{figure}
  \centering
  \includegraphics[width=0.48\linewidth]{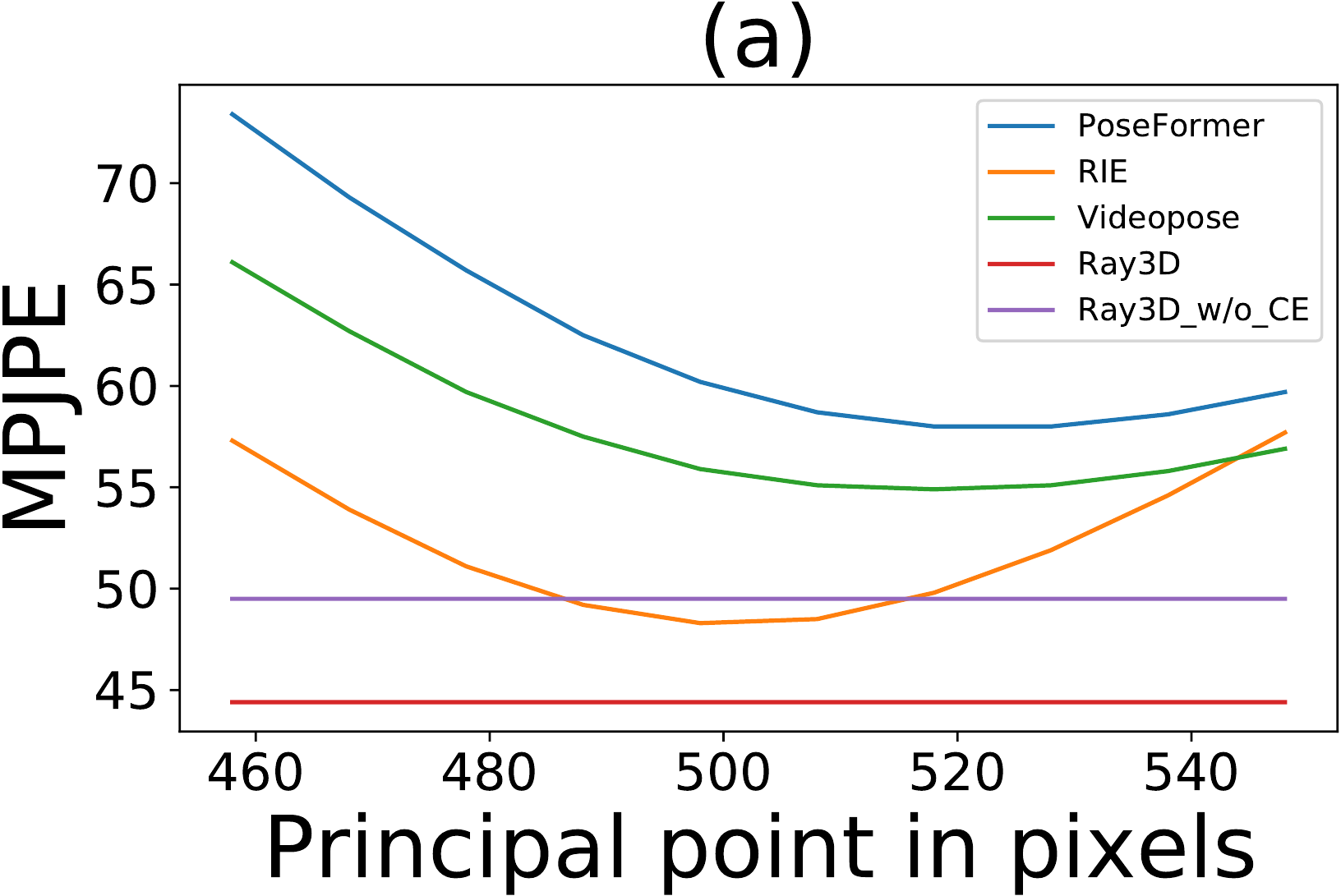}
  \includegraphics[width=0.48\linewidth]{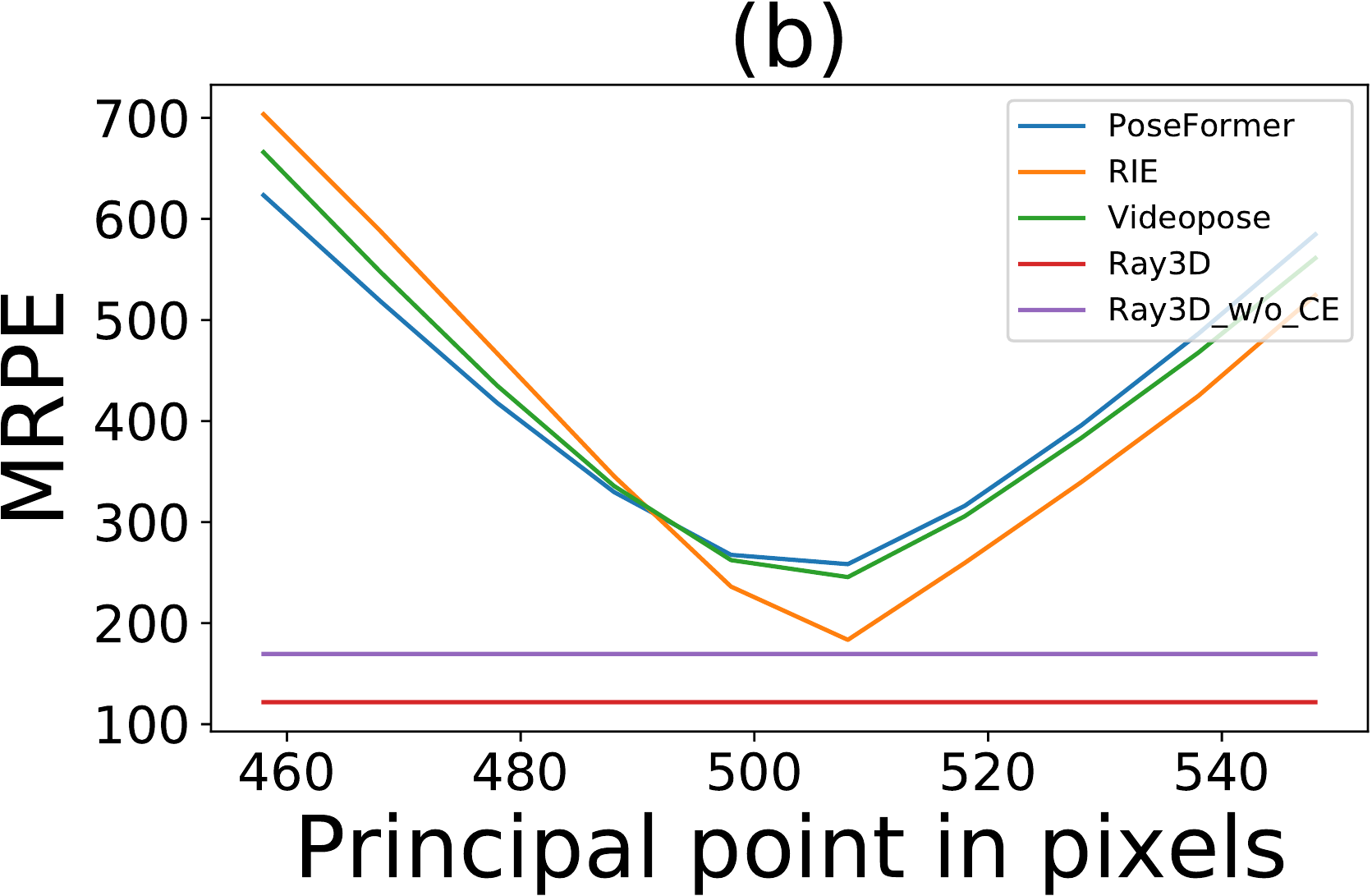}
  \caption{Performance under MPJPE and MRPE in case of principal point changes are plotted in (a) and (b) respectively. The x-axis represents x-coordinate of 2D principal point of the virtual camera in pixels.}
  \label{fig:intrinsic_pp_supp}
\end{figure}

\noindent\textbf{Intrinsic generalization}
As shown in the Fig.~\ref{fig:intrinsic_pp_supp} (a) and (b), principal point changes affect VideoPose, PoseFormer, RIE to varying degrees under MPJPE and MRPE metrics respectively. In contrast, both Ray3D and Ray3D\_w/o\_CE achieve stable result. This result clearly showcases the merits of our ray-based input representation. 

\section{Qualitative results in WCS}
In this section, we provide qualitative results generated by Ray3D and other state-of-the-arts on H36M and 3DHP datasets. Specifically, we visualize 3D keypoints in WCS generated by corresponding methods.

\begin{figure*}
  \centering
  \includegraphics[width=0.98\linewidth]{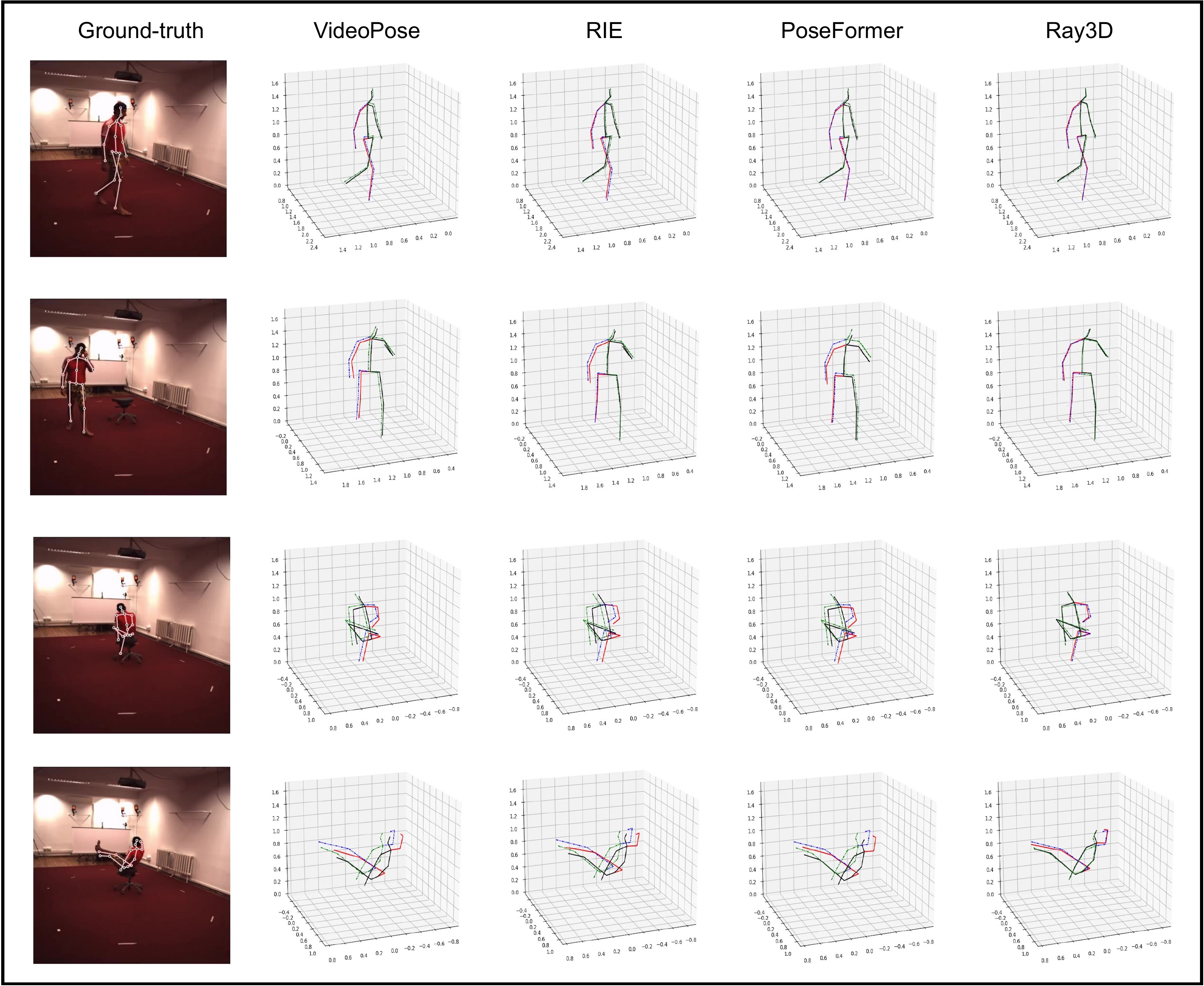}
  \caption{Qualitative comparison of Ray3D with VideoPose, RIE and PoseFormer on H36M. All four models are trained on H36M. First column shows 2D ground-truth poses. Black color denotes left part of person limbs, red color denotes right part of person limbs. 3D estimation results predicted by VideoPose, RIE, PoseFormer and Ray3D are shown in the second, third, forth and fifth column respectively. Dashed lines denote 3D ground-truth poses. Solid lines represent the poses estimated by corresponding approaches. Green and black color lines denotes left part of person limbs, blue and red lines denote right part of person limbs. 17-joint skeleton is visualised.}
  \label{fig:viz_01}
\end{figure*}

\begin{figure*}
  \centering
  \includegraphics[width=0.98\linewidth]{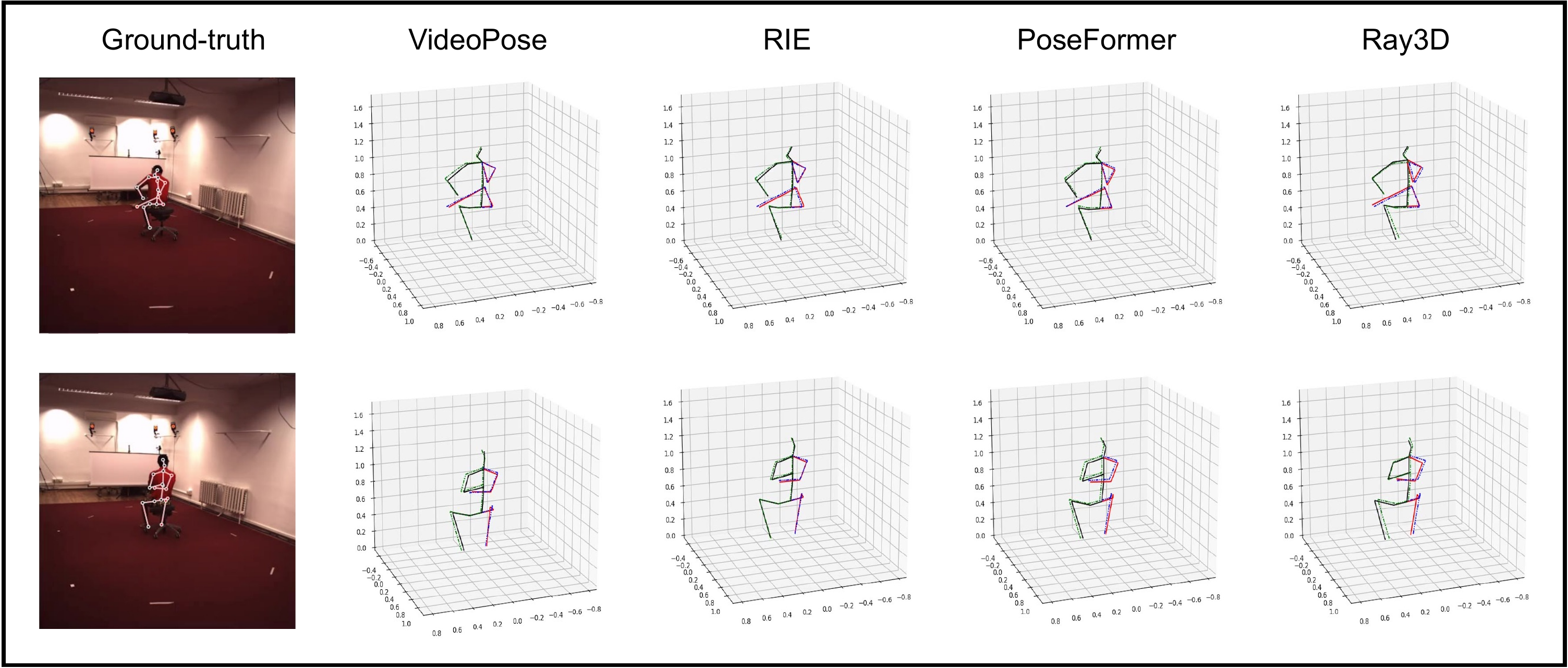}
  \caption{Visualization of inferior performance of Ray3D, compared with other state-of-the-arts on H36M. All four models are trained on H36M. First column shows 2D ground-truth poses. Black color denotes left part of person limbs, red color denotes right part of person limbs. 3D estimation results predicted by VideoPose, RIE, PoseFormer and Ray3D are shown in the second, third, forth and fifth column respectively. Dashed lines denote 3D ground-truth poses. Solid lines represent the poses estimated by corresponding approaches. Green and black color lines denotes left part of person limbs, blue and red lines denote right part of person limbs. 17-joint skeleton is visualised.}
  \label{fig:viz_02}
\end{figure*}

\noindent\textbf{H36M}
Fig.~\ref{fig:viz_01} shows qualitative comparison of Ray3D with VideoPose, RIE and PoseFormer on H36M. We train all four models on H36M with 17-joint definition. From the visualization, we can observe that Ray3D has superior ability to generate more precise location of root joint with comparable root-relative pose estimation results. Fig.~\ref{fig:viz_02} presents two examples of inferior estimation of Ray3D compared to baseline, yet the error is close among these methods. 

\begin{figure*}
  \centering
  \includegraphics[width=0.98\linewidth]{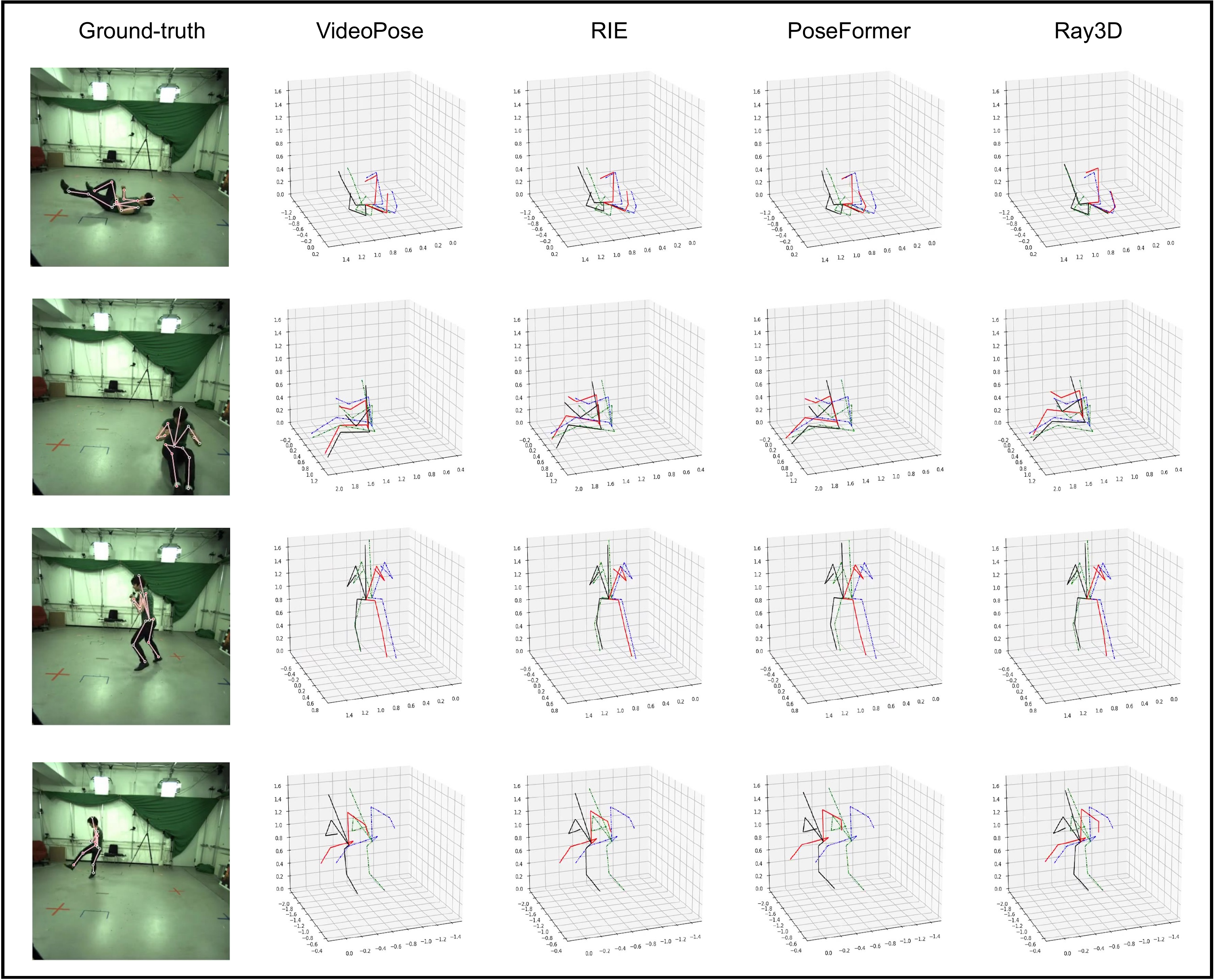}
  \caption{Qualitative comparison of Ray3D with VideoPose, RIE and PoseFormer on 3DHP. All four models are trained on 3DHP. First column shows 2D ground-truth poses. Black color denotes left part of person limbs, red color denotes right part of person limbs. 3D estimation results predicted by VideoPose, RIE, PoseFormer and Ray3D are shown in the second, third, forth and fifth column respectively. Dashed lines denote 3D ground-truth poses. Solid lines represent the poses estimated by corresponding approaches. Green and black color lines denotes left part of person limbs, blue and red lines denote right part of person limbs. 14-joint skeleton is visualised.}
  \label{fig:viz_03}
\end{figure*}

\noindent\textbf{3DHP}
In Fig.~\ref{fig:viz_03}, we present the qualitative comparison of Ray3D with VideoPose, RIE and PoseFormer on 3DHP as well. The models are trained with 14-joint definition. Our Ray3D shows better performance than other state-of-the-arts clearly.

\begin{figure*}
  \centering
  \includegraphics[width=0.98\linewidth]{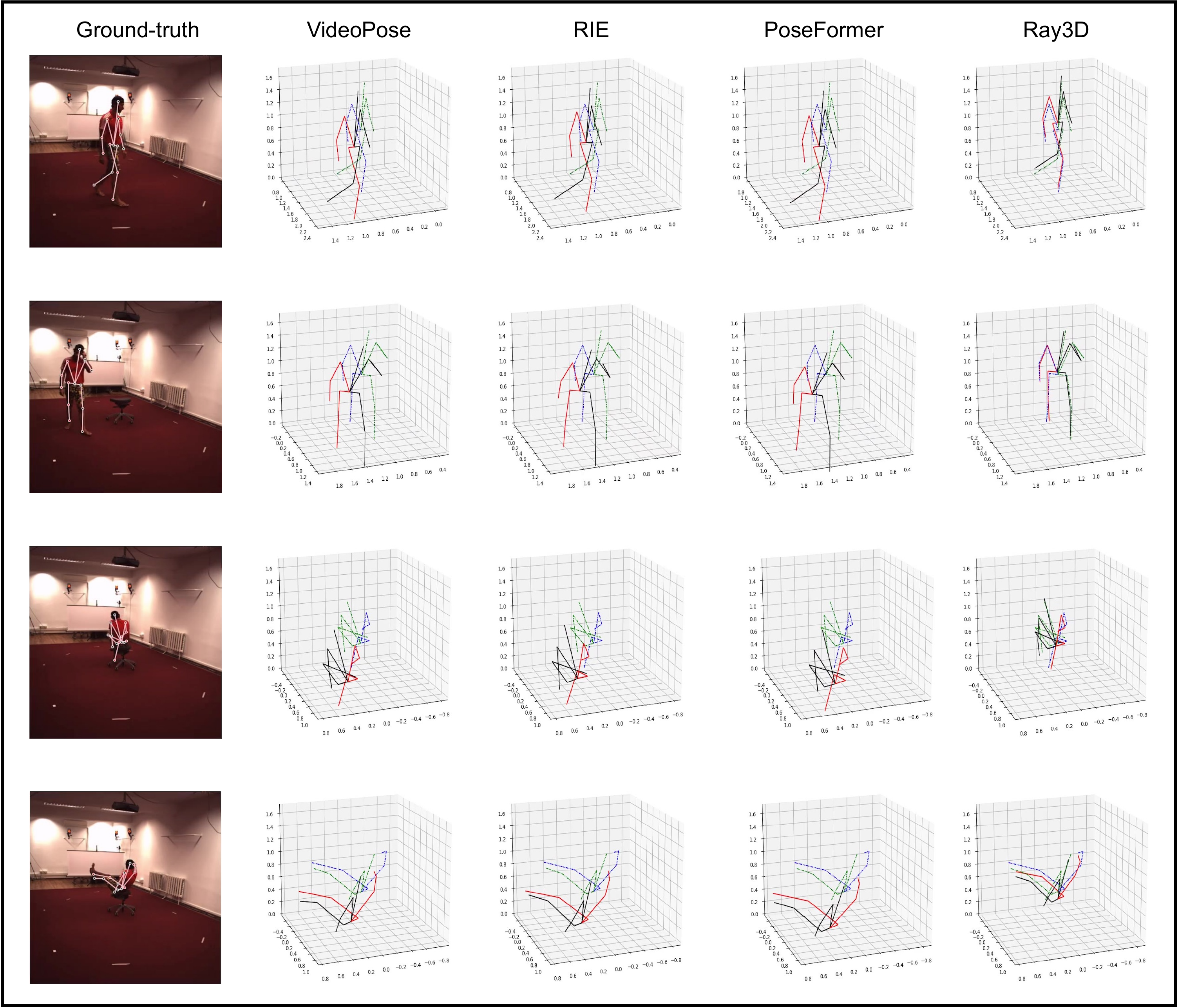}
  \caption{Visualization of generalization of Ray3D, compared with other state-of-the-arts on H36M. All four models are trained on 3DHP. First column shows 2D ground-truth poses. Black color denotes left part of person limbs, red color denotes right part of person limbs. 3D estimation results predicted by VideoPose, RIE, PoseFormer and Ray3D are shown in the second, third, forth and fifth column respectively. Dashed lines denote 3D ground-truth poses. Solid lines represent the poses estimated by corresponding approaches. Green and black color lines denotes left part of person limbs, blue and red lines denote right part of person limbs. 14-joint skeleton is visualised.}
  \label{fig:viz_04}
\end{figure*}

\noindent\textbf{Cross-dataset}
In Fig.~\ref{fig:viz_04}, we compare generalization of Ray3D with other state-of-the-arts on H36M. We train all four models on 3DHP and test them on H36M with 14-joint definition. One can clearly observe that Ray3D generates more accurate estimation results than other approaches, benefiting from our normalized ray representation and camera embedding design.

\clearpage

\end{document}